\documentclass[10pt,twocolumn,letterpaper]{article}

\usepackage[algorithms]{wacv}      

\usepackage{graphicx}
\usepackage{amsmath}
\usepackage{amssymb}
\usepackage{booktabs}
\usepackage{wrapfig}
\usepackage{siunitx}
\usepackage{multirow, multicol}

\usepackage{balance}

\usepackage{makecell}
\usepackage{pifont}

\usepackage[pagebackref,breaklinks,colorlinks]{hyperref}

\usepackage[capitalize]{cleveref}
\crefname{section}{Sec.}{Secs.}
\Crefname{section}{Section}{Sections}
\Crefname{table}{Table}{Tables}
\crefname{table}{Tab.}{Tabs.}

\def\wacvPaperID{*****} 
\def\confName{WACV}
\def\confYear{2025}
\usepackage[dvipsnames]{xcolor}
\usepackage{multirow}
\usepackage{soul}

\colorlet{my_gray}{gray!10}

\begin{document}


\title{\textit{AdaOcc}: Adaptive-Resolution Occupancy Prediction}

\author{
    Chao Chen$^{1}$ \,
    Ruoyu Wang$^{1,2}$ \,
    Yuliang Guo$^{2}$ \,
    Cheng Zhao$^{2}$ \,
    Xinyu Huang$^{2}$ \,
    Chen Feng$^{1}$ \,
    Liu Ren$^{2}$\thanks{The corresponding author: Liu Ren {\tt\small liu.ren@us.bosch.com}}\\
    $^{1}$New York University \quad $^{2}$ BOSCH Research North America\\
    \href{https://ai4ce.github.io/DeepMapping2/}{\tt\small https://github.com/ai4ce/Bosch-NYU-OccupancyNet/}
}
\maketitle

\providecommand{\titlevariable}{AdaOcc}




\maketitle
\begin{figure*}[ht]
    \centering
    \includegraphics[width=0.85\textwidth]{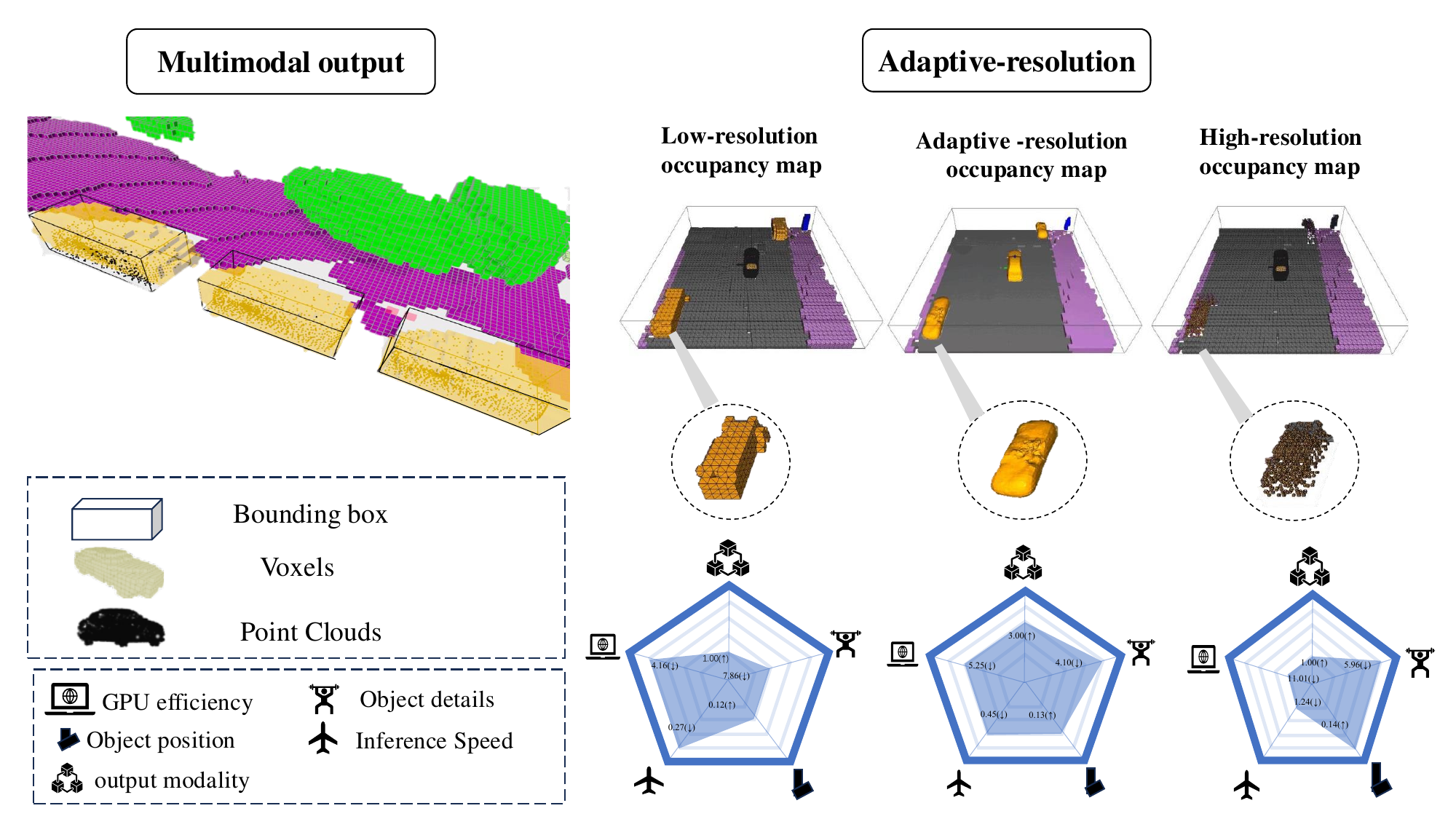}
    \caption{\textbf{\titlevariable} is a multimodal, adaptive-resolution approach designed for high precision in regions of interest while maintaining overall resource efficiency. On the left, \titlevariable's outputs are visualized using bounding boxes, voxels, and point clouds. On the right, visual comparisons are made between low, adaptive, and high-resolution occupancy map predictions, evaluated across five indicators that assess both accuracy and efficiency. The adaptive-resolution approach represents the data in surface format by applying surface reconstruction to point clouds. It yields the most balanced scores, effectively accommodating diverse driving tasks while managing computational costs.}
    \vspace{-2mm}
    \label{fig:MM-HR}
\end{figure*}

\begin{abstract}
Autonomous driving in complex urban scenarios requires 3D perception to be both comprehensive and precise.
%
Traditional 3D perception methods focus on object detection, resulting in sparse representations that lack environmental detail. Recent approaches estimate 3D occupancy around vehicles for a more comprehensive scene representation. However, dense 3D occupancy prediction increases computational demands, challenging the balance between efficiency and resolution.
High-resolution occupancy grids offer accuracy but demand substantial computational resources, while low-resolution grids are efficient but lack detail. To address this dilemma, we introduce \textit{\titlevariable}, a novel adaptive-resolution, multi-modal prediction approach.
Our method integrates object-centric 3D reconstruction and holistic occupancy prediction within a single framework, performing highly detailed and precise 3D reconstruction only in regions of interest (ROIs). These high-detailed 3D surfaces are represented in point clouds, thus their precision is not constrained by the predefined grid resolution of the occupancy map.
We conducted comprehensive experiments on the nuScenes dataset, demonstrating significant improvements over existing methods. In close-range scenarios, we surpass previous baselines by over 13\% in IOU, and over 40\% in Hausdorff distance. In summary, \textit{\titlevariable} offers a more versatile and effective framework for delivering accurate 3D semantic occupancy prediction across diverse driving scenarios.

\end{abstract}
\section{Introduction}
\label{sec:intro}
\vspace{-2mm}


Accurate representation of surroundings is vital for autonomous driving decision-making. The required perception granularity varies based on tasks: highways may need sparse but long-range views, while urban areas require dense, close-range detail. Finding a representation enabling safe navigation in all scenarios, adapting to dynamic changes in real-time, remains a significant challenge.

Various scene representations have emerged in autonomous driving research. The prototypical ones include uniform voxel-based representations~\cite{wang2023openoccupancy, wei2023surroundocc, zhang2023occformer, cao2022monoscene, li2023voxformer}, bounding box representations~\cite{huang2021bevdet, xie2022m, wang2023panoocc, tong2023scene, jiang2023polarformer, wang2023exploring}, implicit representations~\cite{sucar2021imap, rosinol2023nerf, zhu2022nice}, point-based representations~\cite{keller2013real, schops2019bad, whelan2015elasticfusion}, and other forms of representations~\cite{matsuki2023gaussian, mukasa20173d}. While object-centric representations using bounding boxes were traditionally popular, voxel-based representations have become more prevalent in recent years due to the rich information they offer in 3D semantic occupancy maps. Since each voxel contains both occupancy and semantic information, voxel-based representations provide a more comprehensive understanding of the scene. They include additional background descriptions and capture surface shapes with a certain level of granularity. Moreover, voxel-based representations are popular for their seamless integration with navigation and planning frameworks.

\begin{figure*} [h]
    \centering
    \includegraphics[width=1.0\textwidth]{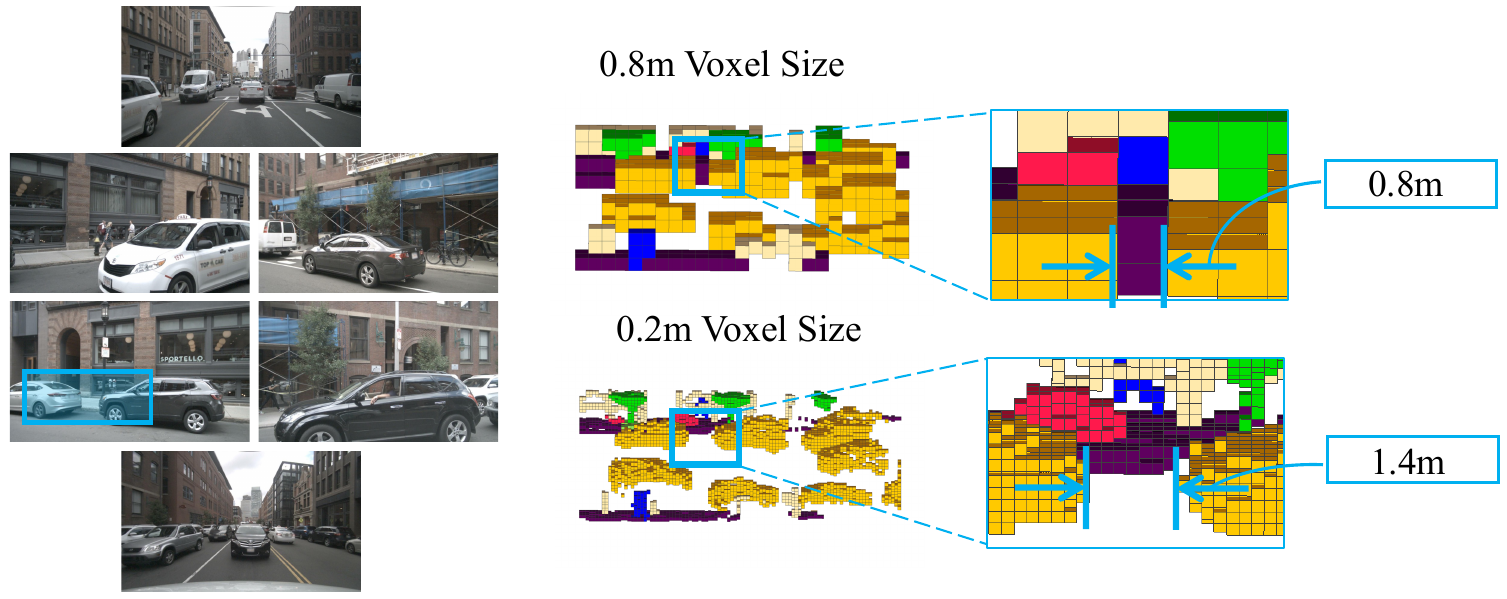}
    \caption{\textbf{Impact of Grid Resolution.} The surrounding scene captured by the RGB images can be represented in occupancy maps with different voxel sizes. The measured distance between the same two cars with grid sizes of 0.8m and 0.2m can differ by up to 0.6m (1.4m - 0.8m). This discrepancy can significantly affect precise navigation in safety-critical scenarios.}
    \label{fig:teaser_figure}

\end{figure*}





Despite the flexibility of 3D semantic occupancy maps in adjusting grid sizes, most of the existing methods produce relatively low resolution (\SI{0.4}{\meter} or \SI{0.5}{\meter} sized) occupancy grid \cite{wei2023surroundocc, tong2023scene, tian2024occ3d}, limiting their applications to highway driving. In urban driving or parking scenarios, higher-resolution representation is essential for precise vehicle maneuvering. As seen in Figure~\ref{fig:teaser_figure}, the distance measured between the two cars can deviate as much as 0.6m when using 0.8m and 0.2m voxels. However, increasing resolution leads to a cubic increase in computational complexity and GPU memory consumption. 



To balance memory efficiency and perception accuracy, we propose two strategies:

\textbf{Non-uniform resolution.} In vehicle path planning, close-range elements are considered more important than far-range elements, and objects (e.g., cars, pedestrians) are prioritized over background elements (e.g., roads, sidewalks). Therefore, we propose a non-uniform resolution representation, focusing high-resolution prediction on close-range objects.

\textbf{Multi-modal 3D representation.} This design is based on the two major drawbacks of voxel grid representation. Firstly, its accuracy is limited by voxel size, leading to GPU resource overload when aiming for finer granularity. Secondly, real-world scene sparsity often results in many unoccupied voxels, causing inefficient memory usage.
To address these issues, we propose a multimodal approach for 3D representation, incorporating outputs such as voxel grids, point clouds, and bounding boxes, rather than relying solely on voxel grids for 3D semantic occupancy (as shown on the left of~\cref{fig:MM-HR}). Among these, point clouds are particularly advantageous due to their high detail and independence from voxel size. They do not necessarily require increased memory usage and only indicate the presence of occupied elements, thus overcoming the limitations of voxel grids.


By employing the aforementioned strategies, we introduce \titlevariable, a multi-modal, adaptive-resolution semantic occupancy prediction approach. \titlevariable~prioritizes precision in critical areas vital for autonomous driving decisions, i.e., close-range objects. Rather than uniformly allocating computational resources, \titlevariable~employs fine-detailed reconstruction via point cloud for close-range objects while using coarser 3D voxel grid predictions for 3D semantic occupancy as supplement. We draw inspiration from object-centric methods~\cite{wang2022detr3d, li2022bevformer, wang2023panoocc} that incorporate an Object Proposal Network (OPN), to identify the regions of interest (ROIs) in 3D space. Subsequently, we connect a 3D point cloud decoder~\cite{yang2018foldingnet} to generate detailed point clouds within these ROIs. 
The right panel of ~\cref{fig:MM-HR} demonstrates that \titlevariable's adaptive-resolution representation strikes a balance of high accuracy and efficiency for autonomous driving tasks. 

To enhance the synergy between different modalities, we jointly train a shared backbone (2D bird's-eye view(BEV)~\cite{li2022bevformer} or 3D feature volume ~\cite{wang2023openoccupancy}), providing voxel grids, bounding boxes and point clouds in a unified network architecture. Leveraging multi-modal representation, our occupancy prediction model can be effectively trained using only coarse ground-truth occupancy data plus raw Lidar points while still being evaluated as high resolution occupancy. \titlevariable~ is experimentally compared to previous methods on nuScenes dataset in both close-range and long-range scenarios to demonstrate its effectiveness. Notably, the remarkable improvement in IOU (>13 \%) and Hausdorff distance (> 40\%) metric for close-range evaluations demonstrates our method's superior performance in capturing the details of object reconstruction and achieving precise object position estimation.

In summary, 
our contributions are listed as follows:





\begin{enumerate}
  \item We propose a multi-modal adaptive-resolution method, offering three output representations with high precision in critical regions while maintaining efficiency for real-time applications. 
  \item We develop an effective joint training paradigm that boosts the synergy between the occupancy prediction and object folding branches.
  \item Our approach demonstrates superior accuracy on the nuScenes dataset, particularly excelling in close-range scenarios that require precise maneuvering.
\end{enumerate}

\section{Related Works}


\noindent\textbf{3D semantic occupancy prediction.} 3D semantic occupancy prediction is rapidly evolving, playing a crucial role in achieving precise perception necessary for safely navigating urban environments. Several pioneering works \cite{cao2022monoscene, li2023voxformer} are designed for single-view image inputs, laying the foundation for dense geometry and semantic inference from a singular perspective. In contrast, other methods \cite{huang2023tri, wei2023surroundocc, zhang2023occformer, miao2023occdepth} utilize surround-view images to achieve a comprehensive 360-degree understanding of the environment. Among these, OpenOccupancy \cite{wang2023openoccupancy} provides a benchmark to evaluate occupancy prediction at the finest level, using a 0.2-meter voxel size. 
The proposed method, CONet, was the first to practically realize occupancy prediction at a 0.2-meter voxel scale through a cascaded approach.

Since these approaches rely on uniformly sampled voxels, their precision is largely constrained by the total number of voxels a computing unit can afford. By focusing computational resources on target objects, ~\titlevariable~ achieves highly precise perception in critical regions while maintaining the overall computation cost.

\vspace{2mm}

\noindent\textbf{3D object detection from surround-view images.} The landscape of camera-based surround-view 3D object detection in autonomous driving has seen significant advancements in unified framework design, as demonstrated by~\cite{liu2022petr, huang2021bevdet, xie2022m, zhang2023simple}. 
Researchers have concentrated on transforming multiple perspective views into a unified 3D space within a single frame, as demonstrated by studies such as~\cite{wang2022detr3d, liu2022petr, huang2021bevdet, li2022bevformer, jiang2023polarformer, wang2023object}. This process can be categorized into two main approaches: (1)BEV-based methods\cite{huang2021bevdet,li2022bevformer,xie2022m,huang2023fast,li2023bevdepth,jiang2023polarformer,wang2023panoocc,tong2023scene,kumar2024seabird}, (2) sparse query based methods~\cite{wang2022detr3d, liu2022petr,lin2022sparse4d,chen2022polar,wang2023object}. In comparison, BEV methods are considered more compatible with other 3D perception tasks demanding dense outputs, such as 3D occupancy prediction, depth estimation, and 3D scene reconstruction.

Inspired by \cite{wang2023panoocc, tong2023scene}, \titlevariable~further enhances 3D comprehension by integrating object detection, occupancy prediction, and object surface reconstruction into a unified framework. Our framework not only comprehensively represents the entire scene but also focuses on high surface precision within object regions. While \cite{wang2023panoocc} employs a similar strategy by performing semantic occupancy prediction within certain ROIs, it still outputs 3D voxels in uniform grids. This approach continues to face the efficiency-precision dilemma in choosing the grid resolution, as seen in other occupancy prediction methods.



\vspace{2mm}
\noindent\textbf{2D-3D encoding backbones.} 
Within the realm of 2D-3D encoding backbones, two predominant methodologies emerge: transformer-based backbones \cite{li2022bevformer, luo2022detr4d, liu2023petrv2, park2022time} and Lift-Splat-Shoot (LSS)-based backbones \cite{philion2020lift, wang2023openoccupancy, li2023voxformer}. Transformer-based backbones typically create a query grid in 3D space, project these grid points onto 2D image planes, and then aggregate the extracted features back to the query grid using a deformable transformer \cite{zhu2020deformable}. Conversely, LSS-based backbones incorporate a depth probability prediction module that allocates 2D image features across 3D space according to estimated depth probabilities. Each approach offers distinct advantages. For our experiments, we chose BEVFormer \cite{li2022bevformer} (transformer-based) and CONet \cite{wang2023openoccupancy} (LSS-based) as the baseline networks to underscore the capabilities of \titlevariable.

\vspace{2mm}

\noindent\textbf{Multi-resolution 3D representations.} Multi-resolution representations have gained considerable traction across various fields in computer graphics and geometric modeling, as evidenced by seminal works~\cite{kobbelt1998interactive,guskov1999multiresolution,de2002multiresolution,michikawa2001multiresolution}. Several approaches~\cite{dai2017shape,hanocka2019meshcnn,jiang2020sdfdiff,wang2018global} adopt a hierarchical method for shape reconstruction, starting with a preliminary low-resolution model that is progressively refined into a high-resolution output. Other methods~\cite{liu2020neural,sun2024behindtheveil} extend hierarchical structures, such as octrees of implicit functions, to represent the radiance field for neural rendering, yet the granularity of the octree is predetermined by the depth map input
In contrast, both MDIF~\cite{chen2021multiresolution} and FoldingNet~\cite{yang2018foldingnet} offer representations of object shapes with adjustable levels of detail.

Within the field of occupancy prediction, CONet~\cite{wang2023openoccupancy} pioneers a coarse-to-fine strategy, refining only the occupied areas in the coarse occupancy map to achieve the first practical 0.2-meter semantic occupancy prediction method. Building solely upon the coarse occupancy map of CONet, our approach significantly improves performance in terms of Hausdorff distance and reduces memory usage, as detailed in Section~\ref{sec:experiment}.

\section{Methodology}

\textbf{Problem statement.} We formulate our task as a multi-modal, adaptive-resolution occupancy prediction. The input of the network is a set of surround-view input images $\mathcal{I} = \{I_1, I_2, ..., I_N\}$, and the output of the network are in \textbf{multiple modalities}, including: (1) 3D semantic occupancy map $M\in\mathbb{R}^{{H}\times {W}\times{D}}$, where this map spans from $X\subset[x_{min}...x_{max}]$, $Y\subset[y_{min}...y_{max}]$, $Z\subset[z_{min}...z_{max}]$. (2) a set of bounding boxes represented by translation$\left(x_i, y_i, z_i\right)$, rotation$\left(q x_i, q y_i, q z_i, q w_i\right)$, sizes $\left(h_i, w_i, d_i\right)$, and (3) object shape in point cloud format $\mathbb{R}^{N\times K\times3}$, here we pick K = $2500$. 

By means of \textbf{adaptive-resolution}, we aim to create a mixed-resolution occupancy map that combine fine resolution for objects with coarse resolution for all matters. The grid size for the occupancy map includes 0.2m, 0.4m, 0.8m, and so on. In this work, we define high resolution as which the voxel size is less than or equal to 0.2m. Otherwise, it is low resolution.

\begin{figure*}[t]
\vspace{-2mm}
    \centering
    \includegraphics[width=0.94\textwidth, trim=0 0 3 0, clip]{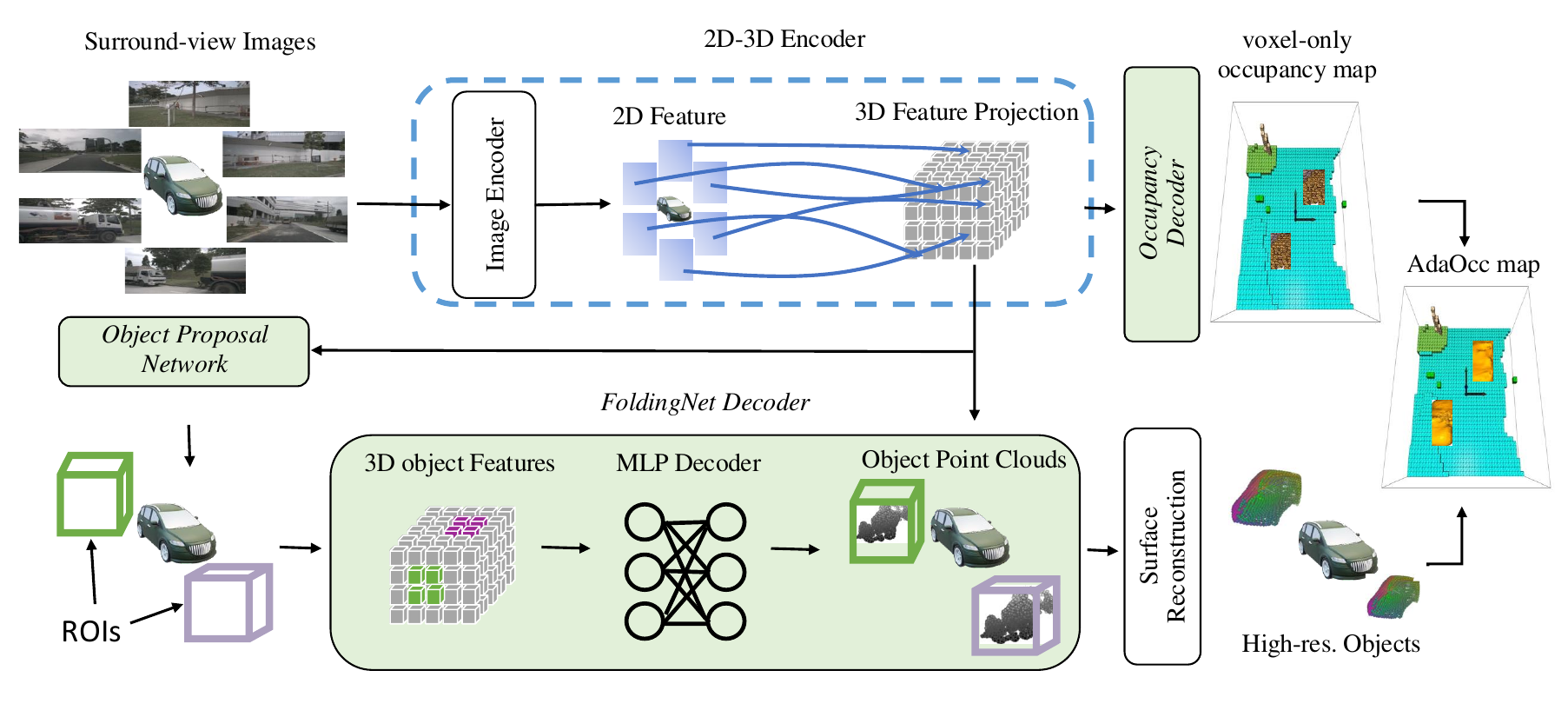} \quad
\vspace{-4mm}
    \caption{\textbf{\titlevariable~Pipeline.} We combine a low-resolution occupancy map and a high-resolution object point cloud to create a adaptive-resolution map. The green boxes represent three outputs from our spatial-temporal encoder: occupancy prediction, 3D object detection, and point cloud reconstruction. Depending on the backbone, the 2D-3D encoder varies. BEVFormer projects image features to a BEV feature volume, while CONET projects them to a 3D feature volume. We consider a BEV feature as a special case of a 3D feature volume where depth dimension equals to 1.}
    \label{fig:workflow}
\vspace{-5mm}
\end{figure*}

\textbf{Architecture overview.} Our approach is versatile, capable of integrating with either BEVFormer~\cite{li2022bevformer} or CONet~\cite{wang2023openoccupancy}, as depicted in Figure~\ref{fig:workflow}. It processes six surround-view input images $\mathcal{I}$ through a 2D-3D encoder. Specifically, images $\mathcal{I}_t$ captured at time $t$ are processed using a CNN to extract 2D image features. These features are subsequently projected into a 3D feature volume $F$ that facilitates semantic occupancy prediction, object detection, and object surface reconstruction. The BEV feature is considered a specific instance of the 3D feature volume.

\subsection{Occupancy Decoder}

Ego-centric occupancy perception is designed to create semantic occupancy maps of fixed grid size in surround-view driving scenarios. This module is intended to provide a holistic understanding of the entire area, allowing for the use of a low-resolution occupancy decoder for greater efficiency. The surrounding occupancy labels, denoted by $V_t$, are predicted through:
\begin{equation}
    V_t=MLP\left(F_{t-1}, F_{t}\right),
\end{equation}
where using BEVFormer as the framework, the previous and the current feature volume ($F_{t-1}, F_t \in \mathbb{R}^{H \times W \times D \times C_{vox}}$) are fed into an MLP-based voxel decoder, to get the coarse 3D semantic occupancy prediction ($V_t \in \mathbb{R}^{H \times W \times D}$), $C_{vox}$ represents feature dimension. All CONet variants(using CONet as backbone) only relies on the current feature ($F_t$) to predict $V_t$ and then applies an additional step of attention-based occupancy refinement on top of the original occupancy prediction.

\subsection{3D Object Detector}\label{sec:OPN}
The 3D object detector is designed to generate 3D object bounding boxes that facilitate object-centric shape reconstruction or can be directly used for downstream tasks. The predicted 3D object bounding box, denoted as $\hat{B}$, is defined as follows:

\begin{equation}
    \hat{B} = \arg\max_{B} P(\text{object}|B) \cdot P(B|I),
\end{equation}
where $P(\text{object}|B)$ is the probability that an object is present given the bounding box 
$B$, $P(\text{object}|B)$ is the likelihood of the bounding box $B$
given the input image $I$. We follow the setup from DETR to regress $900$ bounding boxes and compute their object classification scores. 
The accuracy of the 3D object detector is critical for detailed surface reconstruction, as it directly influences the quality of the results.

\subsection{FoldingNet Decoder}

The FoldingNet decoder processes 3D features and predictions, producing finely detailed surfaces for targeted objects. Initially, FoldingNet~\cite{yang2018foldingnet} utilizes PointNet~\cite{qi2017pointnet} to encode an object's point cloud and then decodes the latent features into another point cloud at an arbitrarily selected resolution. Adapting this process to effectively leverage existing 3D features within our unified framework to directly output highly accurate surface points introduces distinct challenges. Moreover, driving scenarios often involve only partial observations of an object; the variability in partial visibility combined with inaccuracies in the predicted object boxes can further complicate the object-centric decoding process.


\textbf{Box-Aligned Object Feature Aggregation.} Leveraging our object proposal network, we can concentrate on reconstructing the point cloud for each object individually. We establish a regular sampling grid $G = {p_i}, p_i \in \mathbb{R}^3$ within each provided 3D bounding box. The sampling grid is initially created in the object coordinate frame, in alignment with the three dimensions of the bounding box. 
Then $G$ is transformed to the ego-vehicle coordinate frame via the object pose $T$. 
We retrieve the set of feature vectors from the 3D feature volume $F$ using $G$ and apply max pooling over all sampled features to obtain an "object feature vector" $\mathbf{c}$. If features appear at a floating-point location, cubic interpolation is employed to retrieve the feature. The insight behind the sampling process is that \textit{the features within a bounding box encodes the local surface shape related to the object.} The max pooling operation enhances the robustness of $\mathbf{c}$ to errors in bounding box prediction and to issues of partial visibility. The sampling process can be represented using the following equation:
\begin{equation}
    \mathbf{c} = \text{maxPooling}(\{F(T(p_i))\}),\ p_i \in G.
\end{equation}
In training, we use ground truth bounding boxes poses to transform the 3D sampling grid, while in testing we use predicted bounding boxes poses. 

\textcolor{black}{\textbf{Point cloud decoding.} After the feature encoding the object shape is retrieved, the object surface point cloud $\mathbf{P}$ is decoded as:
\begin{equation}
    \mathbf{P} = f_{\theta}(\mathbf{c}, \mathbf{g}),
    \end{equation} 
where $\mathbf{g}$ is the 2D sampling grid used by the FoldingNet decoder $f_{\theta}$, which is a multi-layer perceptron (MLP) to decode the point cloud.}




\vspace{-2mm}
\subsection{Joint training and losses}
\vspace{-2mm}

In our approach, we address the challenge of object-centric occupancy prediction through the integration of three components. A joint training paradigm can effective enhance the synergy between different modules. The effectiveness of the joint training is further validated via ablation study in supplemental materials.

Our joint training approach integrates a combination of losses from various modules: semantic occupancy loss, object detection loss, and surface reconstruction loss. The \textbf{semantic occupancy loss} ($\mathcal{L}_{\mathrm{sem}}$), which utilizes focal loss, is designed for predicting semantic occupancy within a fixed grid size. For object detection, we employ the \textbf{object detection loss} ($\mathcal{L}_{\mathrm{det}}$), incorporating both focal loss for classification and L1 loss for the regression of bounding boxes. This loss function not only selects N valid boxes from a pool of candidates but also accurately estimates the position of each box simultaneously. Furthermore, the \textbf{surface reconstruction loss} ($\mathcal{L}_{\mathrm{surf}}$), using chamfer loss, is applied to the surface reconstruction of foreground objects to ensure precise alignment between the predicted and actual object point clouds. These three loss functions collectively enhance the efficiency of our adaptive occupancy prediction framework. More detailed descriptions of these loss components can be found in the supplementary materials.

\section{Experiment}\label{sec:experiment}
\vspace{-2mm}

We conducted extensive experiments using the NuScenes dataset, which included evaluations of occupancy predictions at both close and full ranges, as well as object detection. Because comparing voxelized ground truth with multi-modal output presents challenges, we converted our detailed object point clouds into an occupancy representation with a grid size of 0.2m for evaluation purpose. This resolution matches the ground truth voxel size in CONet and OpenOccupancy~\cite{wang2023openoccupancy}.
\vspace{-2mm}

\subsection{Dataset}\label{dataset_metric}

Our experiments engage NuScenes dataset to assess our object-centric occupancy prediction methods.
In our experimental configuration, the ground truth labels encompass a bounded range in the x-direction from -50.0 to +50.0 meters, in the y-direction from -50.0 to +50.0 meters, and in the z-direction from -5.0 to 3.0 meters. Furthermore, to evaluate a approach's performance under various voxel resolutions, we partition the space into voxels with granularity settings of 0.2 meters, 0.4 meters, and 0.8 meters. Given above experimental setups, we evaluate candidate methods within the varying defined spatial boundaries and at different voxel resolutions. Note that the ground truth 3D semantic occupancy comes from~\cite{wang2023openoccupancy}. Note that our setup is similar to the OpenOccupancy Benchmark~\cite{wang2023openoccupancy}, except that we retain very small object ground truth (GT) boxes in both our training and validation sets.
\vspace{-2mm}
\subsection{Baselines}\label{baseline}

In this paper, we selected BEVFormer~\cite{li2022bevformer} and CONet~\cite{wang2023openoccupancy} as state-of-the-art baselines for our evaluation. We aim to enhance these baselines with a multi-resolution representation to demonstrate the robustness and flexibility of our approach in improving occupancy prediction accuracy across various methods. BEVFormer, which utilizes object detection for scene representation, proves its efficacy in occupancy prediction tasks~\cite{tong2023scene}. We aim to enhance BEVFormer's accuracy through the adoption of a more granular representation method, specifically FoldingNet. Conversely, CONet employs Depth Net for initial rough occupancy predictions, subsequently refined using a transformer. Although its refinement process is conceptually similar to our approach, it lacks flexibility and does not efficiently utilize GPU resources, as it refines all occupancy grids uniformly. Our method focuses on refining predictions particularly for nearby objects, optimizing the overall resource expenditure.

\vspace{-2mm}

\begin{figure*}[t]
    \centering
    \begin{subfigure}{0.32\textwidth}
        \includegraphics[width=\textwidth]{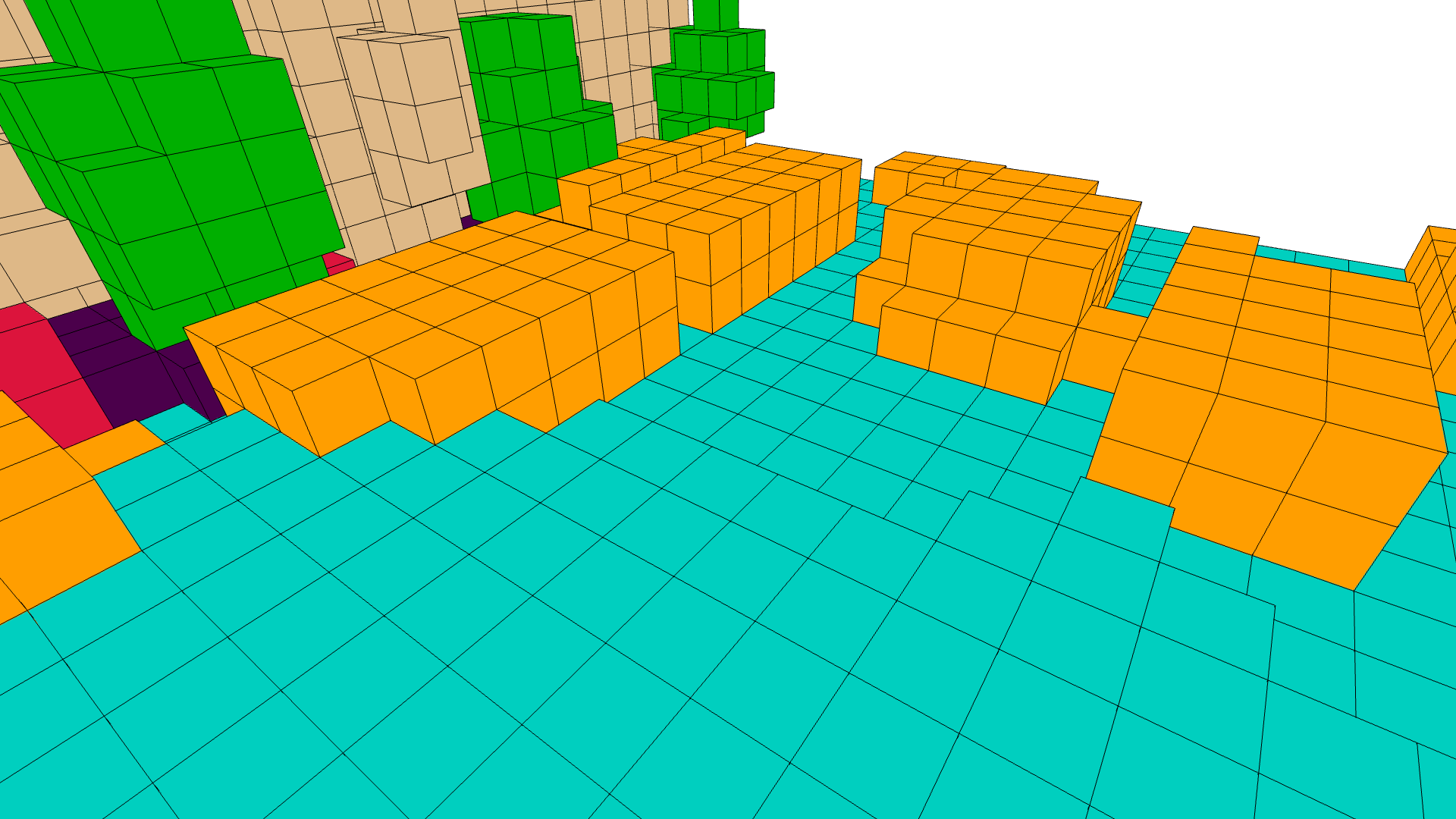}
        \subcaption[]{\textbf{BEVFormer}}
    \end{subfigure}
    \hspace{0.01\textwidth} 
    \begin{subfigure}{0.32\textwidth}
        \includegraphics[width=\textwidth]{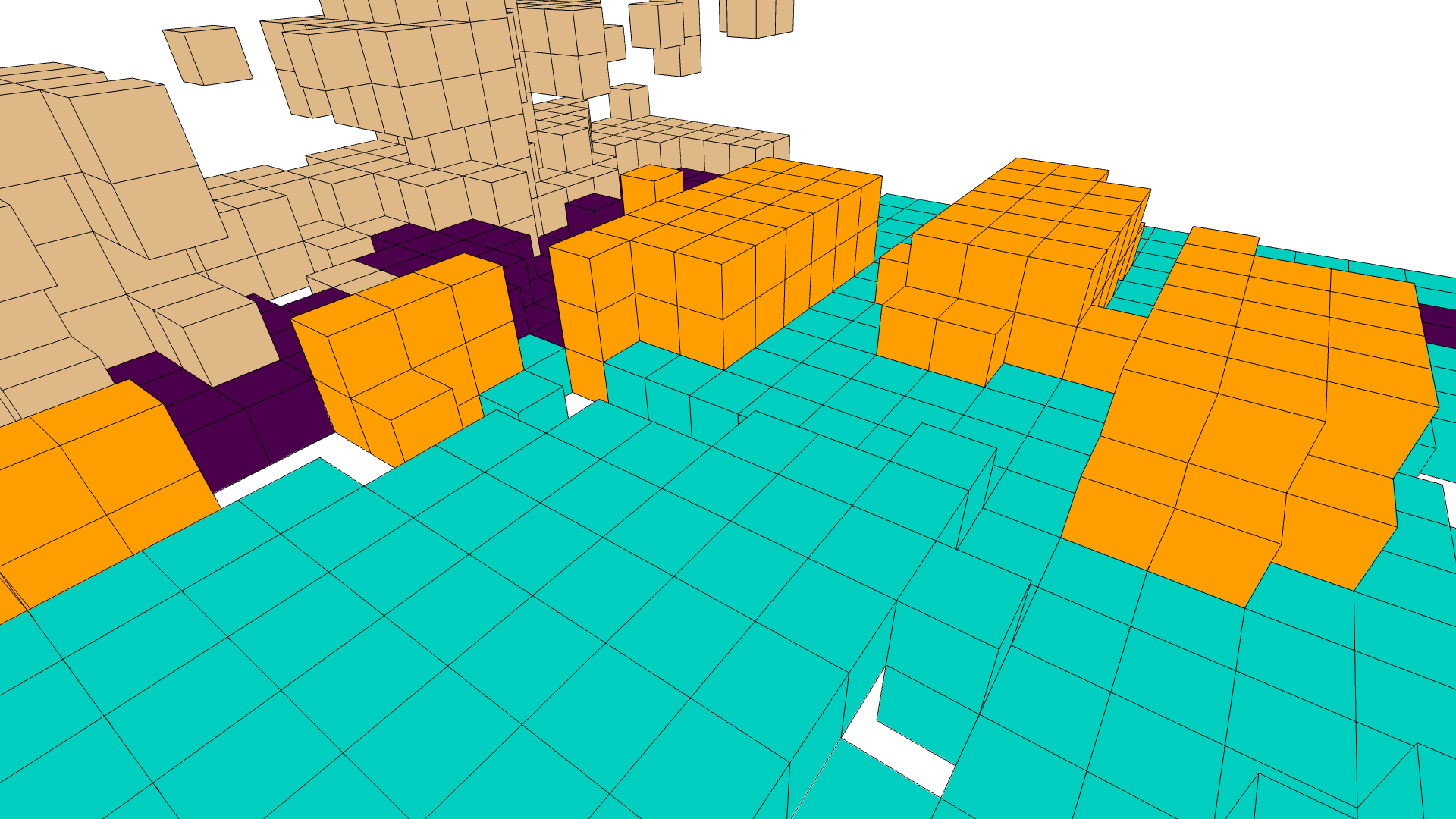}
        \subcaption[]{\textbf{CONet}}
    \end{subfigure}
    \hspace{0.01\textwidth} 
    \begin{subfigure}{0.32\textwidth}
        \includegraphics[width=\textwidth]{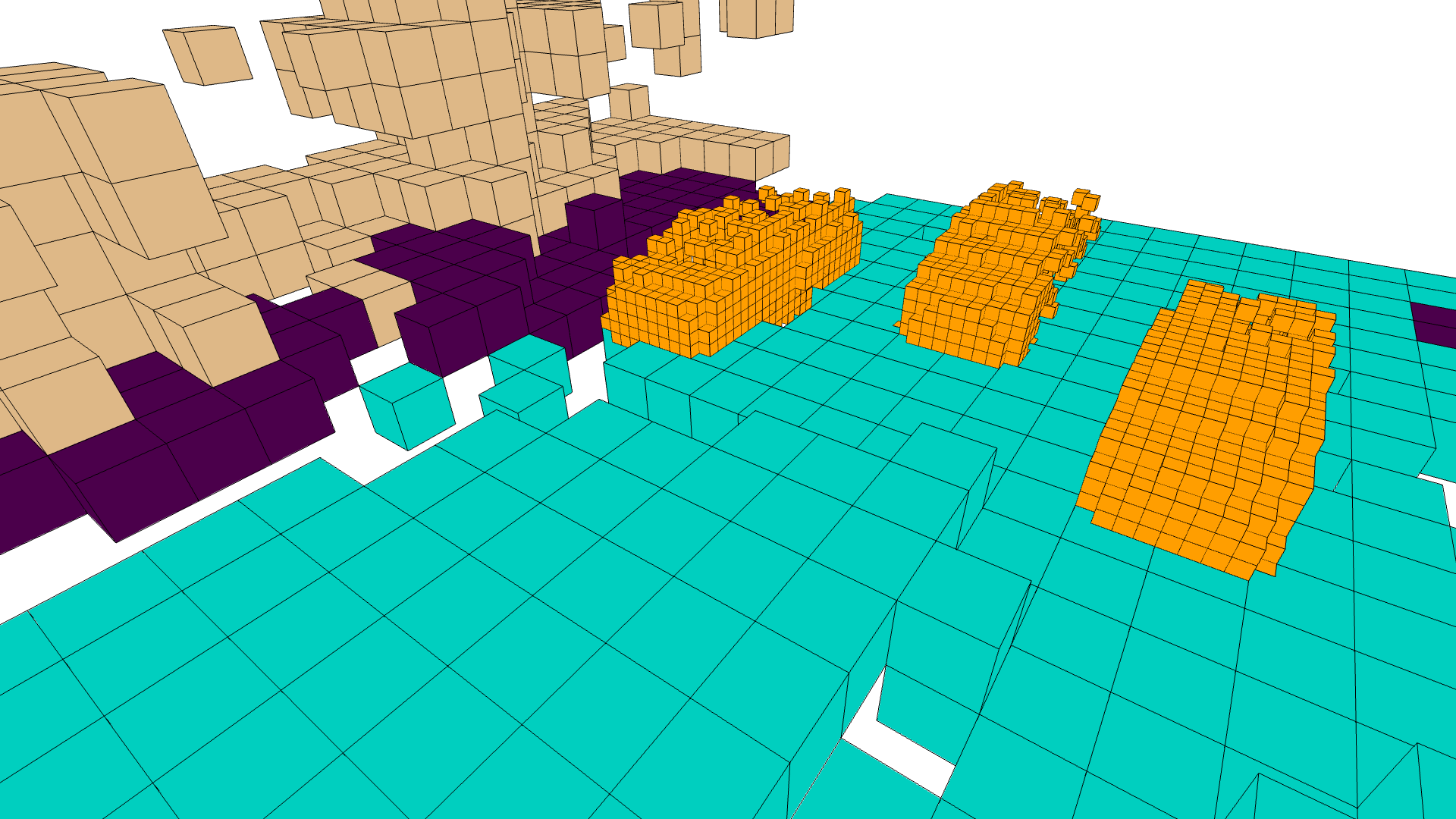}
        \subcaption[]{\textbf{\titlevariable\_B}}
    \end{subfigure}
    \caption{\textbf{Qualitative Results on BEVFormer, CONet, and \titlevariable\_B}. As observed, BEVFormer and CONet involve erroneous connections between different objects. Benefiting from adaptive-resolution, \titlevariable~distinctly separates each car with clear margins.}
    \label{fig:coarse_fine_images}
    \vspace{-3mm}
\end{figure*}

\subsection{Evaluation metric}\label{sec:eval}

We assess the performance of our object detection, and occupancy prediction approach respectively. For object detection, we follow exact same procedure as OccNet~\cite{tong2023scene}, shown in Table.~\ref{tab:detection}. For occupancy prediction, the overall occupancy is evaluated by Intersection over Union (IOU), and per-class occupancy is evaluated by mean Intersection over Union (mIOU). In addition, we apply \textit{Hausdorff Distance}~\cite{javaheri2020generalized} for detailed assessment of the accuracy of object shapes. It evaluates the similarity between the predicted object point cloud and the ground truth object point cloud by measuring the maximum distance between them after \textcolor{black}{bipartite matching}. Hausdorff distance is expected to more adequately describe the precision of object shapes, not only their positional accuracy. Detailed descriptions of the evaluation metrics are provided in the supplementary materials. For 3D voxel grid, we use the centers of voxels as the points to calculate Hausdorff distance. In our experiments, We calculate the Hausdorff distance against ground truth voxel grid using the output point cloud in AdaOcc, while using the finest voxel grid they can get in the baseline methods.

\begin{table*}[h]
\begin{center}
\caption{\textbf{Close-range occupancy evaluation.} We evaluate occupancy prediction in  Hausdroff Distance, IOU, evaluation time per iteration and GPU usage on four methods: BEVFormer, CONet,~\titlevariable\_B(\titlevariable~based on BEVFormer), and ~\titlevariable\_C (\titlevariable~based on CONet). The close range spans from -12.8 to +12.8 meters in both x and y directions, and from -5.0 to 3.0 meters in the z-direction. Colored numbers indicate the percentage of improvements over baselines.  Each model is trained at different resolutions but evaluated at the same 0.2m grid size.}
 
\resizebox{1.0\linewidth}{!}{%
\begin{tabular}{l|c|c|c|c|c}
\hline
Method    & \multicolumn{1}{l|}{Train Grid Size(m)} & Hausdorff Distance(m)(↓)                                 & IOU(↑)  & Eval. Time(s) & GPU Usage(GB)   \\ \hline
 \multirow{2}{*}{BEVFormer} & 0.4  & 7.868  & 0.125 & 0.443 & 5.186 \\
                            & 0.8  &   \textemdash    & 0.122 & 0.272 & 4.166     \\ \hline
\multirow{3}{*}{CONet}      & 0.2  & 10.816 & 0.243 & 0.383 & 16.310\\
                            & 0.4  &   \textemdash    & 0.192 & 0.367 & 9.476 \\
                            & 0.8  &   \textemdash    & 0.170 & 0.292 & 8.768 \\\hline
\multirow{2}{*}{\titlevariable\_B} & 0.4  & \multicolumn{1}{c|}{\multirow{2}{*}{\textbf{4.099(\textcolor{teal}{+47.9\%})}}} & 0.142(\textcolor{teal}{+13.6\%})  &0.455 & 5.250 \\
                                 & 0.8 & \multicolumn{1}{c|}{} & 0.140(\textcolor{teal}{+14.75\%})  & 0.315 & 4.171\\ \hline

\multirow{3}{*}{\titlevariable\_C} & 0.2   & \multicolumn{1}{c|}{\multirow{3}{*}{\textbf{5.967(\textcolor{teal}{+44.8\%})}}} & 0.246(\textcolor{teal}{+1.2\%})  &1.348 & 18.274\\
 & 0.4                                  & \multicolumn{1}{c|}{} & 0.197(\textcolor{teal}{+2.6\%})  &1.239 & 11.010 \\
 & 0.8                                  & \multicolumn{1}{c|}{} & 0.193(\textcolor{teal}{+13.5\%})  &0.770 & 10.314\\\hline
\end{tabular}
}
\vspace{-8mm}
\label{tab:closerange}
\end{center}
\end{table*}

\subsection{Close-range occupancy prediction}\label{sec:close-range}

This section evaluates close-range occupancy predictions, crucial for narrow path navigation and parking in autonomous driving.


Without lose of generality, we define close-range as spanning from -12.8 to +12.8 meters in both x and y directions, and from -5.0 to 3.0 meters in the z-direction. Depth estimation within 30 meters has been shown to be accurate~\cite{wang2023exploring}, allowing for precise predictions and enhanced model performance in this range. We train the \titlevariable model with two backbones from BEVFormer and CONet, named AdaOcc\_B and AdaOcc\_C, respectively. We also use three grid resolutions (0.2m, 0.4m, and 0.8m) in training and assess all models by upscaling the grid resolution to 0.2m for Intersection over Union (IOU) calculations. The focus in close-range settings is on obstacle avoidance, and therefore, mIOU is not included. Furthermore, IOU results for BEVFormer are only provided at 0.4m and 0.8m training grid sizes due to memory constraints on GPUs like the RTX 3090.

\textbf{Close-range IOU on BEVFormer.} 
\cref{tab:closerange} shows that AdaOcc based on BEVFormer consistently demonstrates an IOU improvement of at least 13\% for train grid size of 0.4m and 0.8m. Additionally, since BEVFormer already includes an object detection framework, incorporating object surface reconstruction does not add significant overhead in either evaluation time or GPU usage. This ensures that using the folding method on BEVFormer provides a lightweight and efficient way to improve close-range IOU.

\textbf{Close-range IOU on CONet.} 
For CONet, AdaOcc still shows some improvement. However, AdaOcc\_C demonstrates significant improvement at a coarse training grid size (0.8m), but its performance is not as good at a fine training grid size (0.2m). This is because CONet's inherent coarse-to-fine refinement mechanism already provides substantial improvement in occupancy prediction. Adding extra object detection and object surface reconstruction at a fine resolution does not significantly enhance the original results. Additionally, CONet is a resource-intensive method as it refines every coarsely occupied cell. While object surface reconstruction brings some benefits, the extra object detection head and foldingnet head make the gains of AdaOcc at training grid sizes of 0.4m and 0.8m not particularly worthwhile.

\textbf{Qualitative analysis}. Figure.~\ref{fig:coarse_fine_images} demonstrates that the classic occupancy prediction methods are likely to connect different objects of the same class which will inevitably influence the close-range path planning performance.
In contrast, the overall reconstruction quality of ~\titlevariable~for each object is remarkably better than other baselines within a given range. The solution to poor object detection is to conduct object detection and reconstruct the object at a close range. More qualitative comparisons are included in the supplemental materials.
\begin{table}[h]
\begin{center}
\caption{\textbf{Full-range occupancy evaluation.} In this context, \textit{full-range} refers to the area where x ranges from -50.0m to 50.0m, y ranges from -50m to 50m, and z ranges from -5m to 3m. Besides the additional mIOU to provide a more comprehensive per-class measure of object IOU, other annotations follow Table.~\ref{tab:closerange}.}
\resizebox{1.0\linewidth}{!}{%
\begin{tabular}{l|c|cc}
\hline
Method    & \multicolumn{1}{l|}{Train Grid Size(m)} & IOU(↑)         & mIOU(↑)         \\ \hline
\multirow{2}{*}{BEVFormer} &   0.4   & 0.122   & 0.072 \\
                           &   0.8   & 0.089   & 0.053 \\ \hline
\multirow{3}{*}{CONet}      & 0.2                                     & 0.156 & 0.095                      \\
     & 0.4                                     & 0.136          & 0.082                 \\
     & 0.8                                     & 0.120          & 0.074              \\ \hline

\multirow{2}{*}{AdaOcc\_B}  & 0.4   & 0.128(\textcolor{teal}{+4.9\%}) & 0.089(\textcolor{teal}{+23.6\%})          \\
 & 0.8                                     & 0.093 (\textcolor{teal}{+4.4\%})& 0.089(\textcolor{teal}{+67.9\%})       \\ \hline
\multirow{3}{*}{AdaOcc\_C}     & 0.2                                     & 0.157(\textcolor{teal}{+0.6\%})  & 0.093 (\textcolor{red}{-2.1\%})           \\
     & 0.4                                     & 0.136 (\textcolor{teal}{+0.0\%})       & 0.085(\textcolor{teal}{+3.6\%})         \\
     & 0.8                                     & 0.122 (\textcolor{teal}{+1.6\%})         & 0.079(\textcolor{teal}{+6.8\%})         
\\ \hline
\end{tabular}
}
\vspace{-8mm}
\label{tab:fullrange}
\end{center}
\end{table}

\textbf{Analysis of Hausdorff distance.} As discussed, \titlevariable~based on all baselines has the best average Hausdorff distance. However, we have observed that the misinterpretation of bounding box positions and categories, especially for small objects like humans and bicycles, can significantly impact the accuracy of close-range object occupancy prediction tasks. Thus, it explains why we generate an occupancy map by integrating the voxelized object occupancy onto the coarse occupancy map, instead of replacing one by another. We believe that voxelized object occupancy is a good reference and complement to the coarse occupancy map, especially when the coarse occupancy map misinterprets the occupied cell as unoccupied.

\begin{table*}[t]
\begin{center}
\caption{\textbf{Full-range per-object-class occupancy evaluation.} Our \titlevariable~ variants are compared with prior state-of-the-art methods in terms of IOU for each object class at a single grid resolution of 0.8m.}
\resizebox{1.0\linewidth}{!}{%
\begin{tabular}{p{1.0cm}|p{2.2cm}|*{10}{p{1.0cm}|}p{1.0cm}}
\hline
\multicolumn{1}{c|}{\rotatebox{90}{Class Name}}  & \multicolumn{1}{|c|}{\rotatebox{90}{Baselines}}  & \multicolumn{1}{|c|}{\rotatebox{90}{Overall}}  & \multicolumn{1}{|c|}{\rotatebox{90}{Barrier}} & \multicolumn{1}{|c|}{\rotatebox{90}{Bicycle}} & \multicolumn{1}{|c|}{\rotatebox{90}{Bus}} & \multicolumn{1}{|c|}{\rotatebox{90}{Car}} & \multicolumn{1}{|c|}{\rotatebox{90}{Construction}} & \multicolumn{1}{|c|}{\rotatebox{90}{Motorcycle}} & \multicolumn{1}{|c|}{\rotatebox{90}{Pedestrian}} & \multicolumn{1}{|c|}{\rotatebox{90}{Traffic Cone}} & \multicolumn{1}{|c|}{\rotatebox{90}{Trailer}} & \multicolumn{1}{|c}{\rotatebox{90}{Truck}} \\ \hline
\multirow{4}{*}{IOU}  & 
BEVFormer & 0.053  & 0.014  & 0.035 & 0.077 & 0.095 &  0.050 & 0.067 & 0.074 & 0.037 & 0.027  & 0.061 \\ 
& CONet & 0.074 & 0.020 & 0.073 & 0.118 & 0.126 & 0.057 & 0.067 & 0.097 & 0.054 & 0.056 & 0.072 \\ \cline{2-13} 

 & AdaOcc\_B & \textbf{0.089}  & \textbf{0.029}  & \textbf{0.082} & \textbf{0.144} & \textbf{0.159} &  \textbf{0.087} & \textbf{0.103} & \textbf{0.117} & \textbf{0.066} & 0.061  & 0.044 \\ 
 & AdaOcc\_C & 0.079 & 0.028 & 0.071 & 0.141 & 0.128 & 0.077 & 0.072 & 0.049 & 0.041 & \textbf{0.105} & \textbf{0.080} \\ \hline
\end{tabular}
}
\vspace{-6mm}
\label{tab:fullclass}
\end{center}
\end{table*}

\textbf{Evaluation time on different voxel grid sizes.} BEVFormer has the shortest evaluation time because it predicts every occupancy grid equally and coarsely. In contrast, CONet requires the most time for evaluation due to its two-stage process. Initially, it computes a coarse occupancy map similar to BEVFormer, and subsequently, it refines each grid for all occupied grids. As CONet itself lacks an object detection pipeline, its cost for object surface reconstruction will be higher than BEVFormer's, as it requires both an object detection and a foldingnet head for object point cloud reconstruction.

\vspace{-2mm}

\subsection{3D object detection} 
\vspace{-2mm}

3D detection task with 3D box regression coarsely regresses the location of the foreground object can be coarsely regressed. In this section, we prove that the joint training of occupancy prediction, 3D detection, and surface reconstruction can improve the detector performance for all three models (BEVNet, VoxNet, and OccNet)~\cite{tong2023scene}, in terms of mAP, NDS, and other parameters. We developed the object detection pipeline from BEVNet and CONet, and the performance of the object detection method is very similar to the other three baseline methods.

\begin{table}[t]
\begin{center}
\vspace{-2mm}
\caption{\textbf{Object detection evaluation.} Since our work is mainly constructed from BEVFormer and CONet respectively, we compare to those 3D detectors involved in the two baselines or share a similar idea.}
\vspace{-2mm}
\resizebox{1.0\linewidth}{!}{%
\begin{tabular}{l|ccccccc}
\hline
Method     & mAP↑  & NDS↑  & mAOE↓ & mAVE↓ & mAAE↓ & mATE↓ & mASE↓ \\ \hline
BEVFormer     & 0.271 & 0.390 & 0.578 & 0.541 & 0.211 & 0.835 & 0.293 \\
VoxNet     & \textbf{0.277} & 0.387 & 0.586 & 0.614 & 0.203 & 0.828 & \textbf{0.285} \\
OccNet     & 0.276 & 0.390 & 0.585 & 0.570 & \textbf{0.190} & 0.842 & \textbf{0.285} \\ \hline
\titlevariable\_B~ &  0.273     &   \textbf{0.391}    &   \textbf{0.577}    &   0.574    &  0.222     &   \textbf{0.808}    &   0.295    \\ 
\titlevariable\_C &  0.272 &   0.390    &   0.579    &   \textbf{0.532}    &  0.209     &   0.833    &   0.291    \\ \hline
\end{tabular}
}
\label{tab:detection}
\vspace{-6mm}

\end{center}
\end{table}

\subsection{Full-range occupancy prediction}\label{sec:full-range} 

The adaptive-resolution approach represents a computationally efficient and flexible strategy that strikes a balance between accuracy and efficiency. It involves generating a full-range adaptive-resolution occupancy map by incorporating close-range voxelized object occupancy onto a full-range coarse occupancy map. Similar to Sec.~\ref{sec:close-range}, we tested \titlevariable~with grid resolution=0.2m, 0.4m, and 0.8m.


\textbf{Full range IOU and mIOU.} 
\cref{tab:fullrange} illustrates that \titlevariable~based on both baselines outperforms their baseline model at grid resolutions of 0.2m, 0.4m, and 0.8m. However, we can observe that globally, \titlevariable~based on BEVFormer demonstrates more significant improvements. This is consistent with the findings in Section \ref{sec:close-range}. CONet already has some refinement at training grid sizes of 0.2m and 0.4m, so additional object surface reconstruction does not yield significant improvements.

\textbf{Per-object-class evaluation in IOU.} In our occupancy prediction evaluation, we prioritize object segments over static scenes, particularly for classes like pedestrians and vehicles, as their movements are unpredictable. Similarly, Table \ref{tab:fullclass} reveals that \titlevariable~surpasses the other baselines for all 10 object classes. 



Through examining mAP and NDS for BEVFormer and ~\titlevariable\_B shown in~\cref{tab:detection}, we establish that \titlevariable~ slightly outperforms BEVFormer, the foundation upon which \titlevariable~is built. This finding demonstrates that the object surface reconstruction task not only enhances the accuracy of occupancy prediction but also enriches the learned features for object detection.

\vspace{-2mm}
\section{Conclusion}\label{sec:conclusion}
In conclusion, our proposed approach offers a multi-modal adaptive-resolution method, providing three output representations with highly precise surfaces in critical regions, while ensuring efficiency for real-time applications. Additionally, we develop an effective joint training paradigm to enhance synergy between the occupancy and folding networks, resulting in improved near-range occupancy prediction performance. Our methods exhibit superior accuracy on the nuScenes dataset, highlighting a focus on detailed surface reconstruction.

\textbf{{Limitation}.}
We observe that the joint training method does not significantly improve the quality of object detection tasks. Further investigations into the interaction between the coarse occupancy prediction and the object surface reconstruction are needed to boost the consistency between different representations. In addition, the efficiency of the unified framework can be further optimized via more advanced parallelized designs.
{\small
\bibliographystyle{ieee_fullname}
\bibliography{egbib}
}

\clearpage
\section*{Appendix}
\renewcommand{\thesection}{\Alph{section}}
\renewcommand{\thefigure}{\Roman{figure}}
\renewcommand{\thetable}{\Roman{table}}

\setcounter{section}{0}
\setcounter{figure}{0}
\setcounter{table}{0}

We provide in this supplementary more ablation studies and additional visualizations on AdaOcc\_B that could not fit in the paper. In particular, we include (1) more details about the loss, (2) more details about the evaluation metrics, (3) the ablation study examining the impact of the number of bounding boxes in occupancy mapping, (4) the ablation study on the number of folded points for each bounding box, and (5) more occupancy visualizations for all baselines.

\section{Loss Details}
\textbf{Semantic occupancy loss.} We applied focal loss as the semantic occupancy loss. Focal loss is a typical classification loss, specialized to tackle problems such as class imbalances and hard data samples. 

\begin{equation}
\mathcal{L}_{\text {sem}}(M) = \sum_{x_{\text {min }}}^{x_{\text {max }}} \sum_{y_{\text {min } }}^{y_{\text {max }}} \sum_{z_{\text {min } }}^{z_{\text {max }}} -\alpha (1 - p(x, y, z))^\beta \log(p(x, y, z)),
\label{eq:focal}
\end{equation}
where p(.) represents the predicted probability of the correct class, $\alpha$ and $\beta$ are hyperparameters to balance well-classified and hard examples.

\textbf{Object detection loss.} We use a similar loss function as in DETR3D~\cite{wang2022detr3d} for the object detection task.
The object detection loss includes a focal loss and a L1 loss, for the classification and regression of the bounding boxes respectively. The focal loss is similar to Eq.~\ref{eq:focal}, minimizing the discrepancy between the predicted bounding boxes classes and ground truth bounding box classes, and L1 loss minimizes the difference between $N$ predicted bounding box parameters $B_{pred}$ and $N$ corresponding ground truth bounding box parameters $B_{GT}$ as:

\begin{equation}
\mathcal{L}_ \text {reg}=\frac{1}{N} \sum_{i=1}^N\left|B_{\mathrm{pred}}-B_{\mathrm{GT}}\right|.
\label{eq:regress}
\end{equation}

\textbf{Surface reconstruction loss.} We use the Chamfer distance~\cite{yang2018foldingnet} as the surface reconstruction loss. It is a geometric distance-based loss function used for measuring the dissimilarity between two point sets. In the context of our work, it quantifies the discrepancy between the reconstructed surface points and the ground truth points. The Chamfer distance $\mathcal{L}_{Chamfer}$ is defined as:
\begin{equation}
\mathcal{L}_{\text {surf}}=\sum_{\mathbf{x} \in \mathbf{X}} \min _{\mathbf{y} \in \mathbf{Y}}\|\mathbf{x}-\mathbf{y}\|_2^2+\sum_{\mathbf{y} \in \mathbf{Y}} \min _{\mathbf{x} \in \mathbf{X}}\|\mathbf{y}-\mathbf{x}\|_2^2,
\end{equation}
where $\mathbf{X} $ represents the reconstructed points, and $\mathbf{Y}$ denotes the ground truth points clouds.

\section{Evaluation metrics details}
\textbf{Intersection over Union (IOU)}: IOU quantifies the overlap between the predicted and ground truth regions. It computes the ratio of the intersection to the union of these regions, providing an indicator of how well the prediction aligns with the actual data.

Mathematically, IOU is defined as:
\begin{equation}
    I O U=\frac{\text { Area of Intersection }}{\text { Area of Union }}
\end{equation}

\textbf{Mean Intersection over Union (mIOU)}: mIOU is the mean value of IOU computed across multiple instances or classes. It provides a holistic measure of the method's accuracy across various categories.

Mathematically, mIOU is defined as:
\begin{equation}
    m I O U=\frac{1}{N} \sum_{i=1}^N I O U_i, 
\end{equation}
where $N$ is the number of instances or classes and $I O U_i$ represents the IOU value for the 
$i$th instance or class.

However, IoU and mIoU are primarily used to measure the accuracy of object detection or segmentation models, quantifying the overlap between two regions. While they can indicate the disparity between the detected object's position and the true object's position, they do not provide detailed information about shape. Therefore, even with a high IoU, it does not guarantee the accuracy of the detected object's shape. 

\begin{figure*}[t]
\vspace{-2mm}
    \centering {\includegraphics[width=0.9\textwidth
    ]{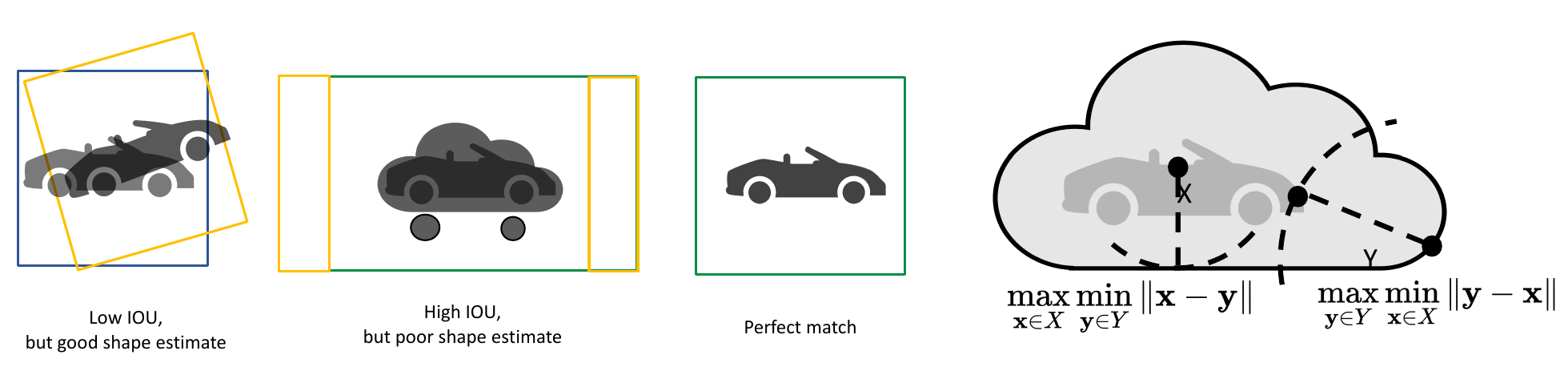}} \quad
    \caption{Misaligned boxes lead to misalignment of the reconstructed objects. Moreover, we also demonstrate how Hausdorff distance works.}
    \label{fig:misaligned}
\vspace{-4mm}
\end{figure*}

\textbf{Hausdorff distance}, on the other hand, offers a deeper metric for assessing the accuracy of object shapes. It evaluates the similarity of the evaluated object point cloud and the ground truth object point cloud by measuring the maximum pairwise distance between them. This implies that Hausdorff distance can better describe the precision of object shapes, not just their positional accuracy. Hence, Hausdorff distance is highly useful in tasks such as shape reconstruction, point cloud matching, etc., where a comprehensive consideration of object shape accuracy is necessary. It is defined as:

\begin{equation}
\mathcal{L}_{\text {Hausdorff }}=\max \left\{ \max_{\mathbf{x} \in X} \min_{\mathbf{y} \in Y} \|\mathbf{x} - \mathbf{y}\|, \max_{\mathbf{y} \in Y} \min_{\mathbf{x} \in X} \|\mathbf{y} - \mathbf{x}\| \right\}
\end{equation}

where $\mathbf{X}$ represents the reconstructed point cloud points, and $\mathbf{Y}$ denotes the ground truth point cloud points. A detailed illustration can be referred to~\cref{fig:misaligned}.

\section{Impact of the number of bounding boxes} We utilize the DeTR head introduced in \cite{li2022bevformer} for the object detection task. The number of bounding boxes is a hyperparameter. We explore this parameter in our ablation study, as presented in ~\cref{tab:topk}.

\begin{table}[h]
\begin{center}
\caption{\textbf{Ablation study on the number of bounding boxes for folding.} We adjust the number of bounding boxes from 0 to 40, with 0 representing the original BEVFormer setup.}
\resizebox{0.8\linewidth}{!}{%
\begin{tabular}{l|c|c|c|c|c}
\hline
\# of box & 10     & 20      & 30    & 40  & 0  \\ \hline
IOU       & \textbf{0.313}  & 0.312   & 0.312  & 0.310 & 0.309 \\
mIOU      & \textbf{0.1590} & 0.1532  & 0.1488 & 0.1475 & 0.1540 \\
Time(s)      & 0.1201 & 0.1298 & 0.1352 & 0.1603 & \textbf{0.1109} \\\hline
\end{tabular}
}
\label{tab:topk}
\end{center}
\vspace{-4mm}
\end{table}

\section{Study on different methods of Compressing 3D into 2D} We investigate various methods of compressing 3D features into 2D, which is required specifically for \titlevariable tasks based on the CONet backbone, such as object detection and folding. The size of the 3D features varies depending on the size of the bounding boxes. Thus, a fixed attention-based weighted aggregation does not work as the input dimension is not fixed. Therefore, we selected four different aggregation methods: max pooling, average pooling, global mean (taking the average over all dimensions), and global max (taking the max over all dimensions). Note that we do not consider summation to be a viable method, as the varying size of 3D features would make the summed 2D features heavily dependent on the size of the 3D features, which is undesirable. It turns out that the max pooling layer performs best, which is similar to the aggregation method used in PointNet, as shown in~\cref{tab:3d2d}.

\begin{table}[h]
\begin{center}
\caption{\textbf{Ablation study on different methods of Compressing 3D into 2D} We investigate various methods of compressing 3D features into 2D, which is required specifically for \titlevariable~ tasks based on the BEVFormer backbone. The methods examined include maxpooling, avgpooling, global-mean, and global-max. In this ablation, we set the train grid size to 0.8m.}
\resizebox{0.8\linewidth}{!}{%
\begin{tabular}{l|c|c|c|c}
\hline
Method & Maxpooling     & Avgpooling      & Global-mean    & Global-max   \\ \hline
IOU       & 0.090  &0.087   & 0.088  & 0.088 \\\hline
mIOU       & 0.053  & 0.051   & 0.049  & 0.047 \\\hline
\end{tabular}
}
\label{tab:3d2d}
\end{center}
\vspace{-4mm}
\end{table}

\section{Study on the number of folded points per Box}
We employ FoldingNet~\cite{yang2018foldingnet} to reconstruct the surface of objects. While the number of folded points for training remains fixed, the number of folded points for testing can be adjusted based on resolution requirements. We conduct an ablation study to analyze how the number of folded points in testing affects IOU and mIOU results, shown in Table.~\ref{tab:foldpoints}.

\begin{table}[h]
\begin{center}
\caption{\textbf{Ablation study on the number of folded points per box.} We vary the number of points from 900 to 40,000. In our previous experiment setup, we opt for 2500 points as it offers the best balance between performance and evaluation time.}
\resizebox{0.7\linewidth}{!}{%
\begin{tabular}{l|c|c|c|c}
\hline
Fold size & 900     & 2500      & 10000    & 40000    \\ \hline
IOU       & 0.312  & \textbf{0.313}   & \textbf{0.313}  & \textbf{0.313}  \\
mIOU      & 0.1530 & 0.1532  & 0.1537 & \textbf{0.1541} \\
Time(s)      & 0.1414 & \textbf{0.14042} & 0.1419 & 0.1444\\\hline
\end{tabular}
}
\label{tab:foldpoints}
\end{center}
\vspace{-5mm}
\end{table}

\section{More occupancy visualization}
Additional occupancy maps for various scenes are displayed. \titlevariable~demonstrates improved separation of the bounding boxes.
\begin{figure*}[ht]
    \centering
    \includegraphics[width=0.32\textwidth]{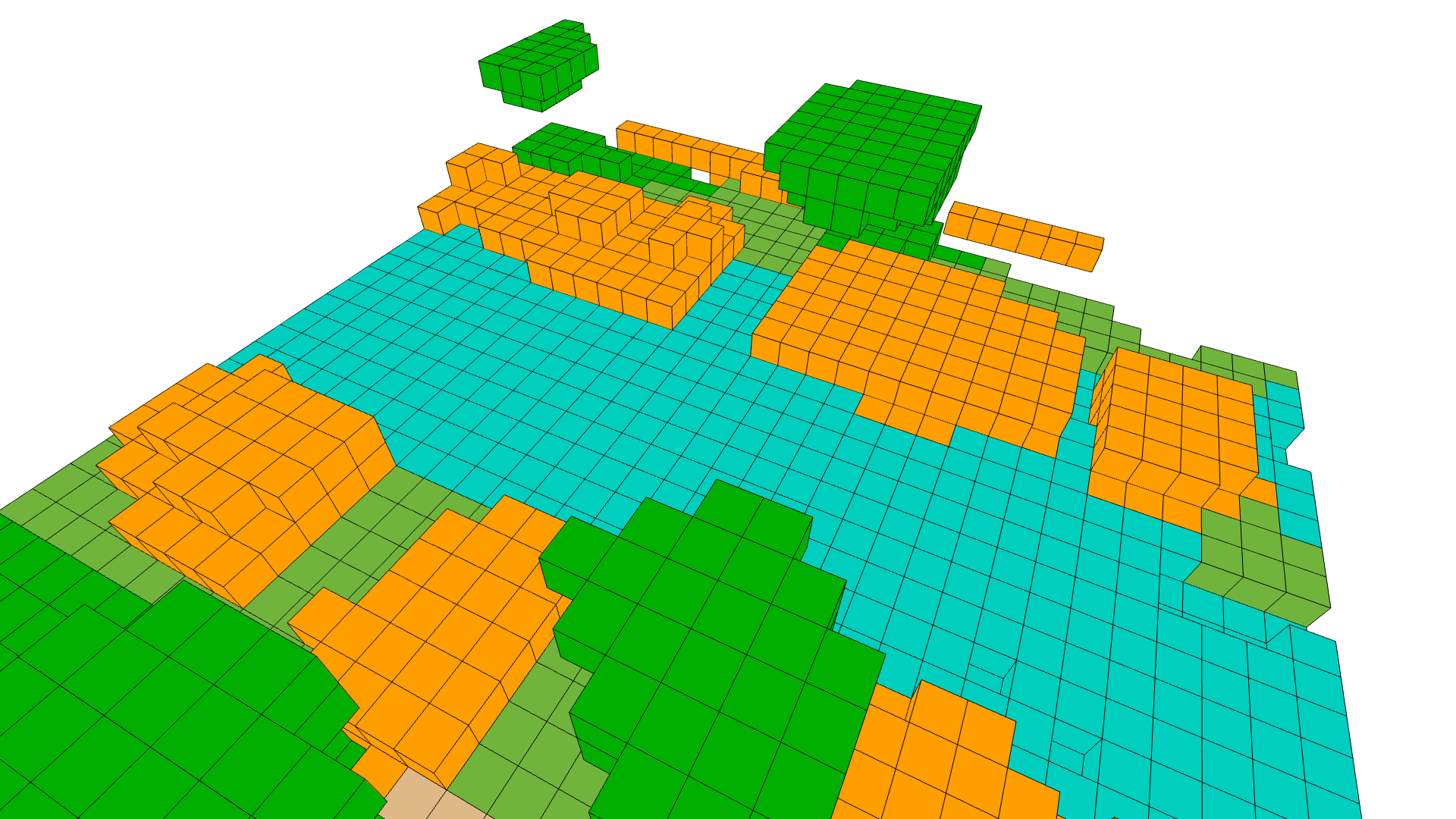}
    \includegraphics[width=0.32\textwidth]{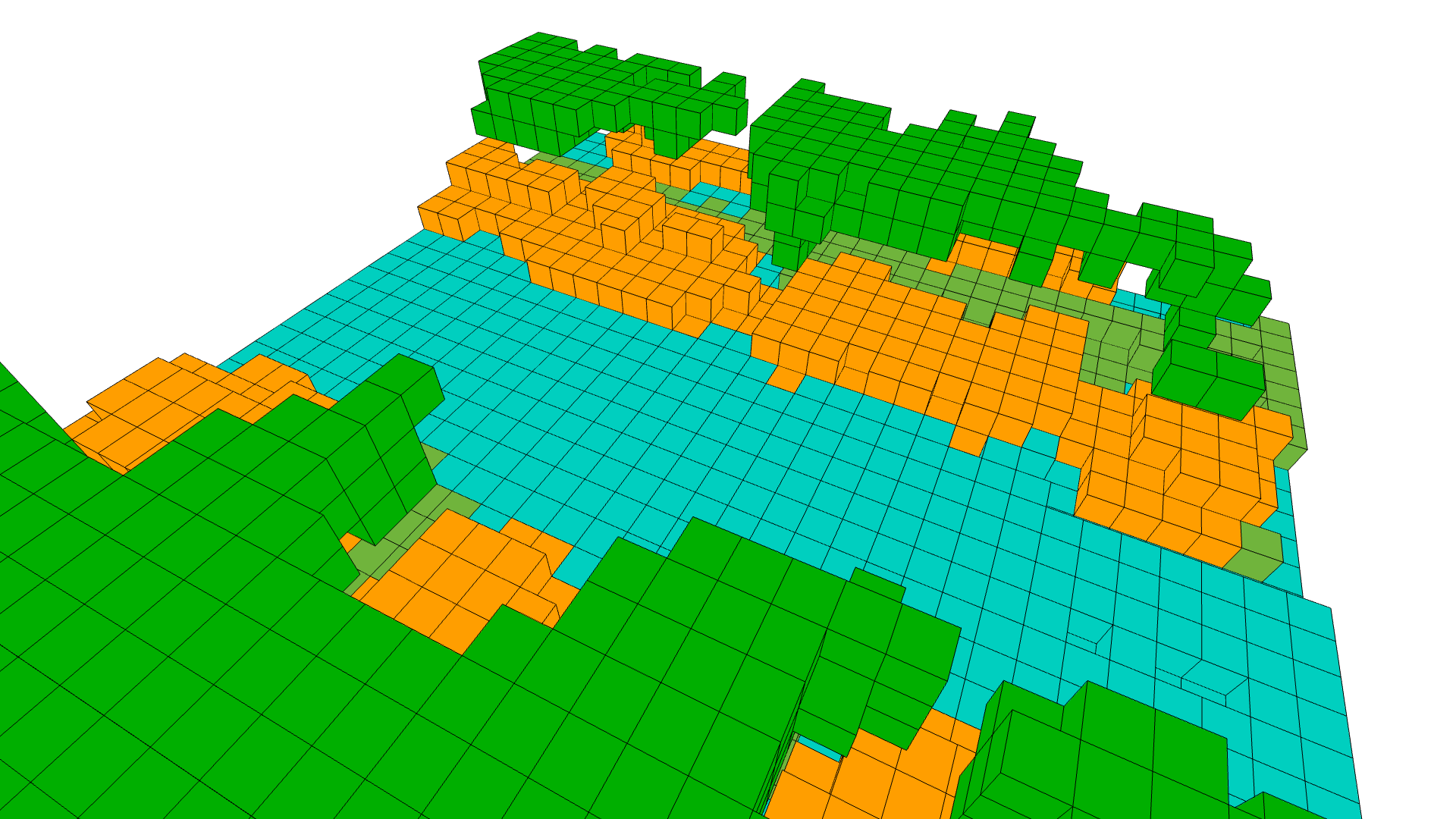}
    \includegraphics[width=0.32\textwidth]{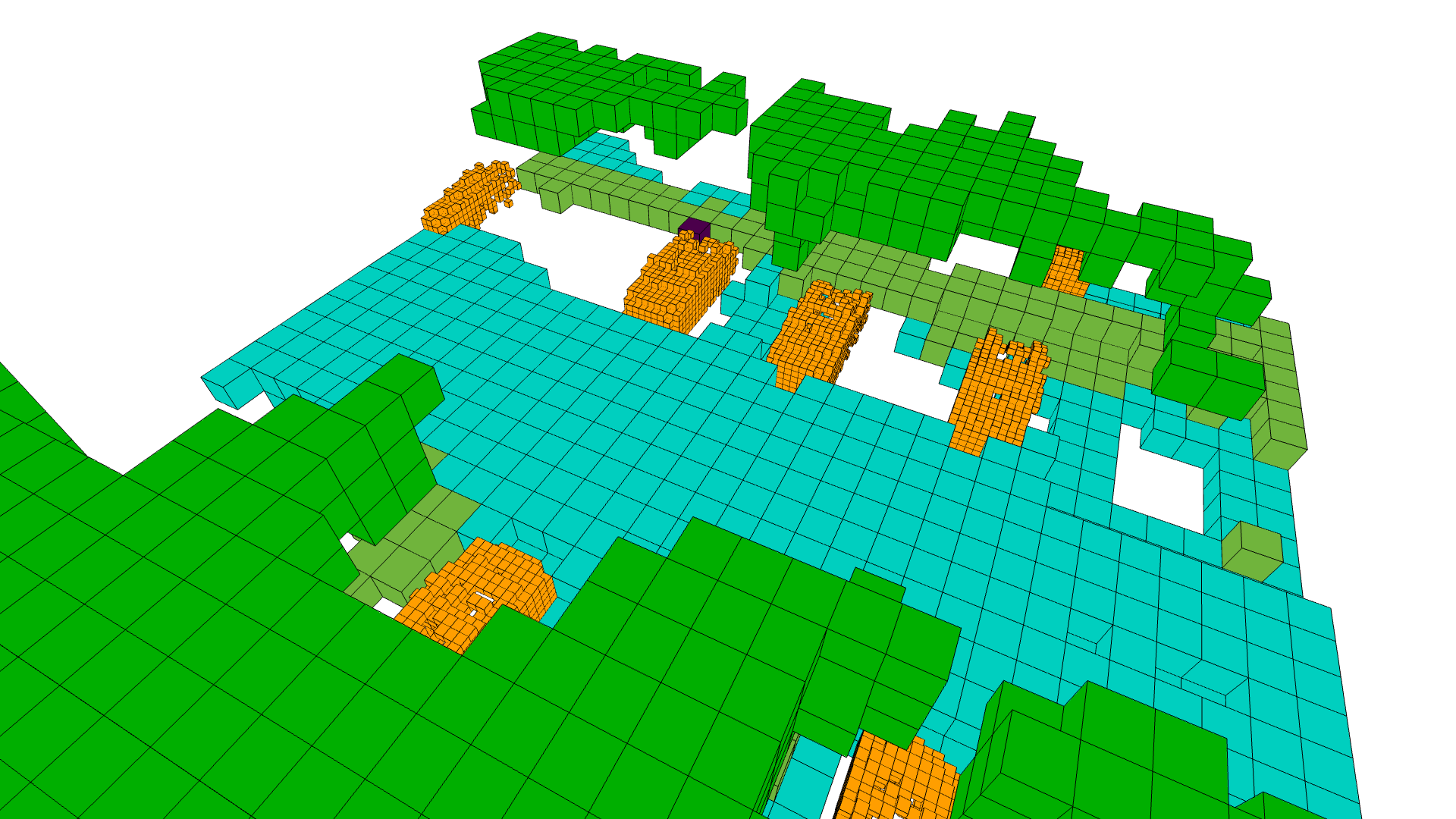}
    \includegraphics[width=0.32\textwidth]{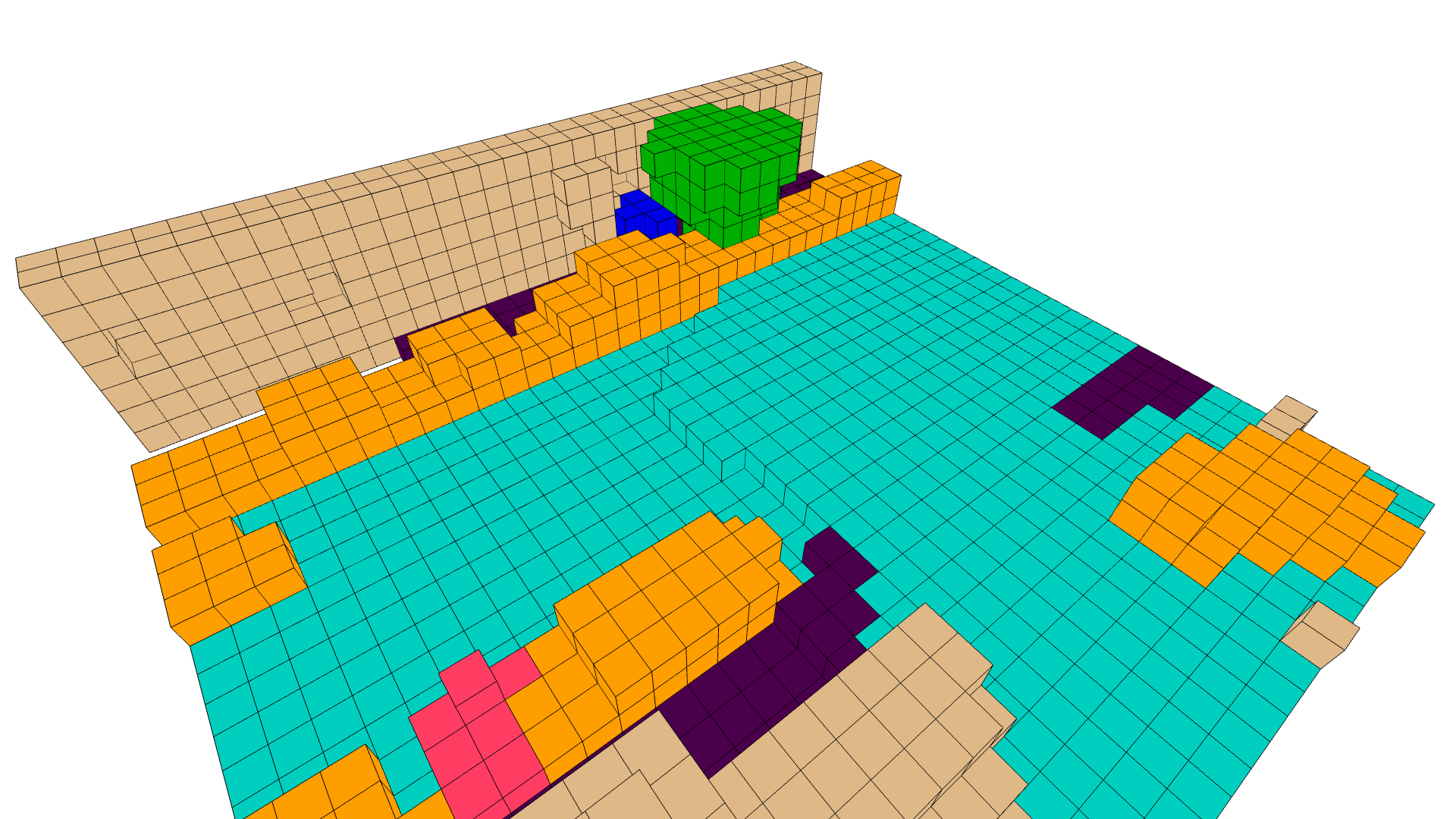}
    \includegraphics[width=0.32\textwidth]{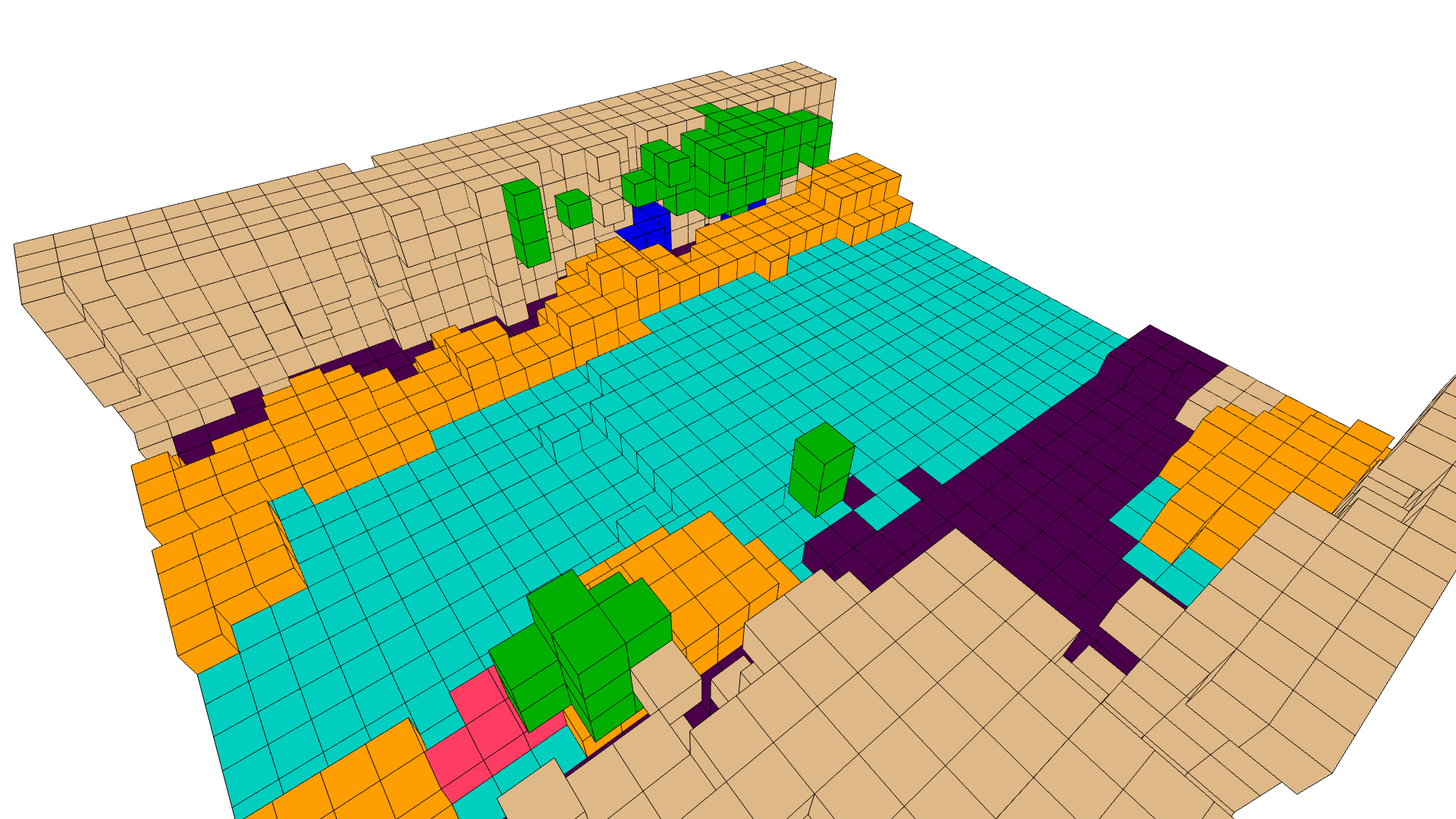}
    \includegraphics[width=0.32\textwidth]{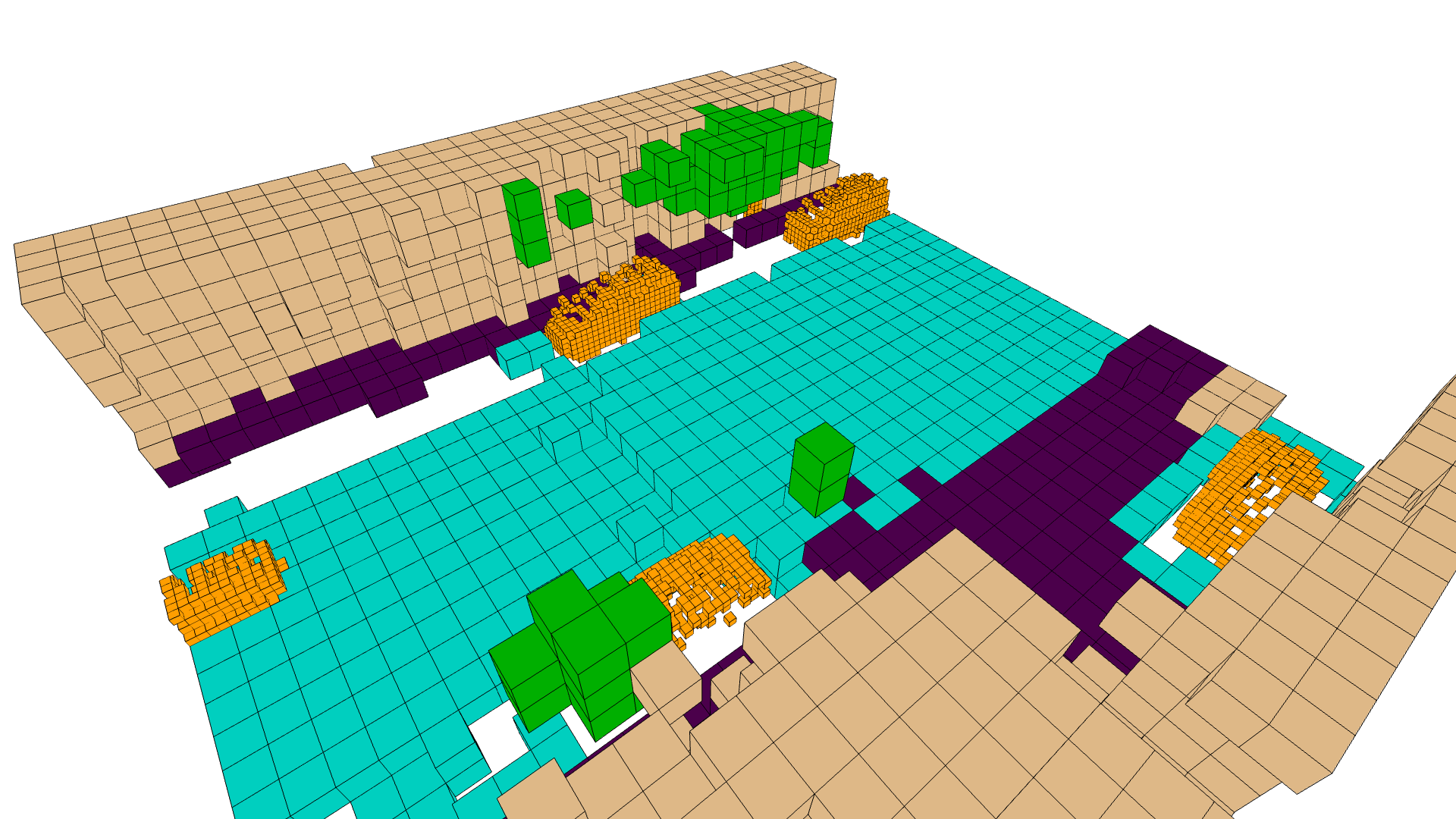}
    \includegraphics[width=0.32\textwidth]{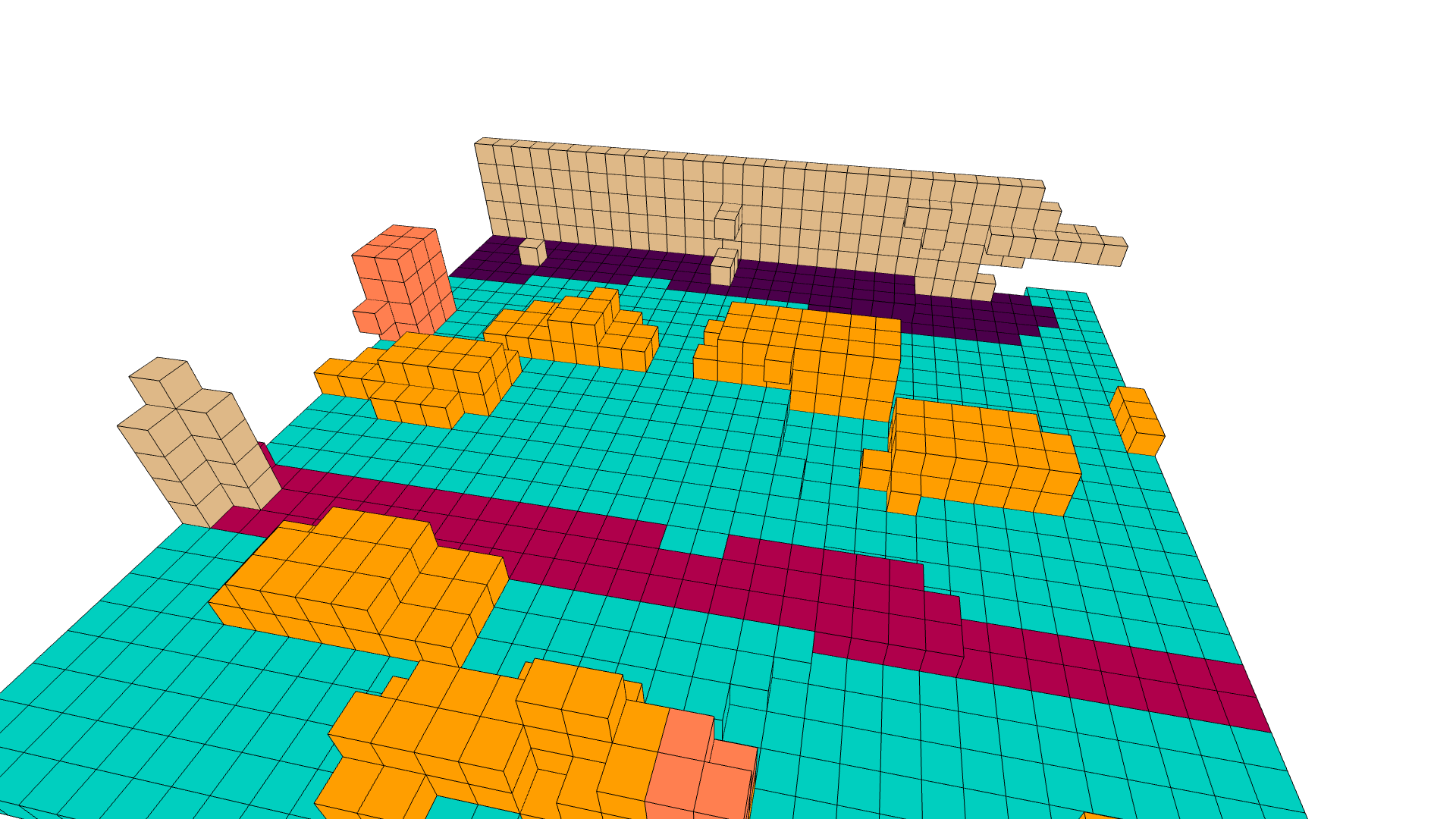}
    \includegraphics[width=0.32\textwidth]{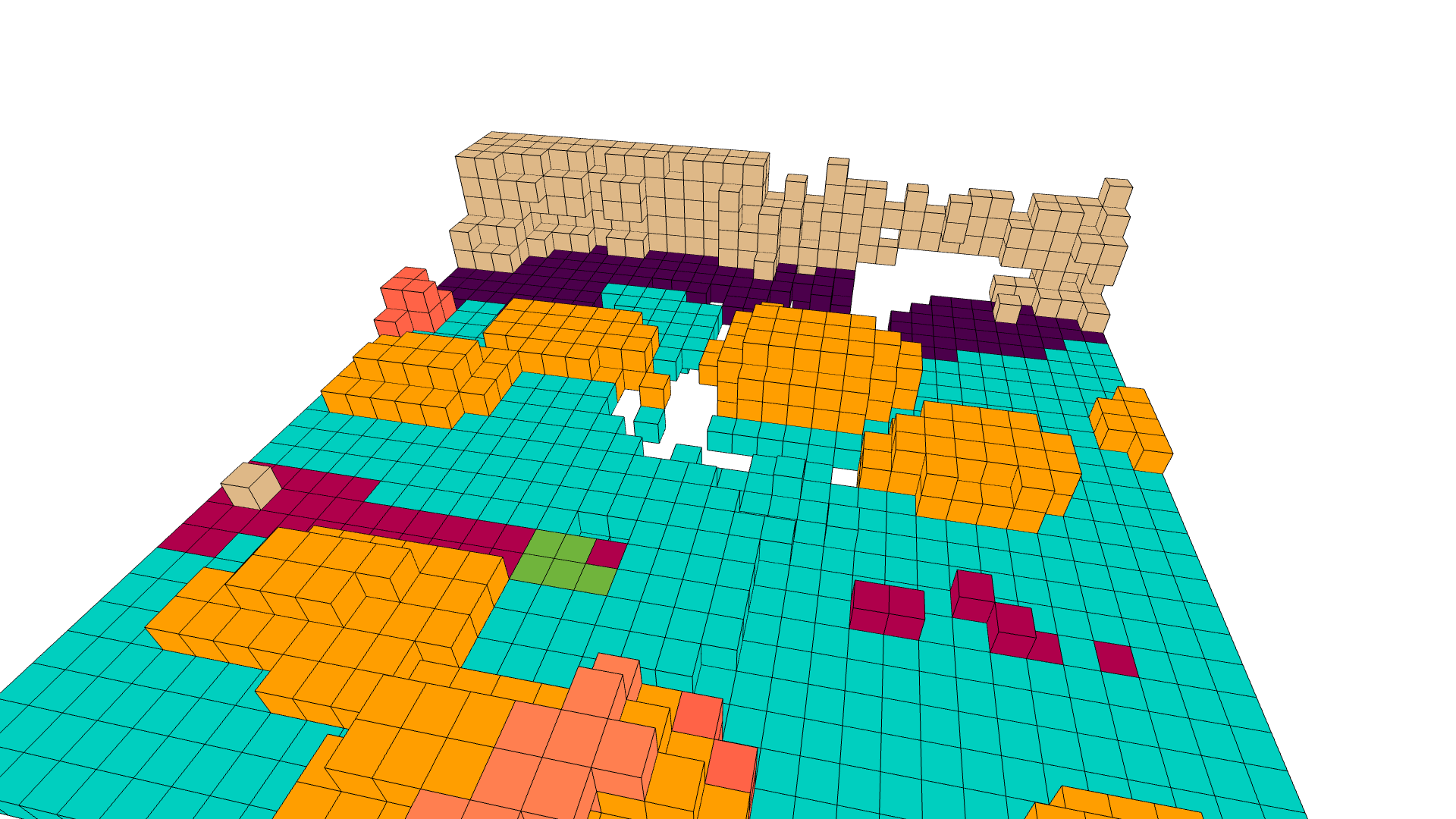}
    \includegraphics[width=0.32\textwidth]{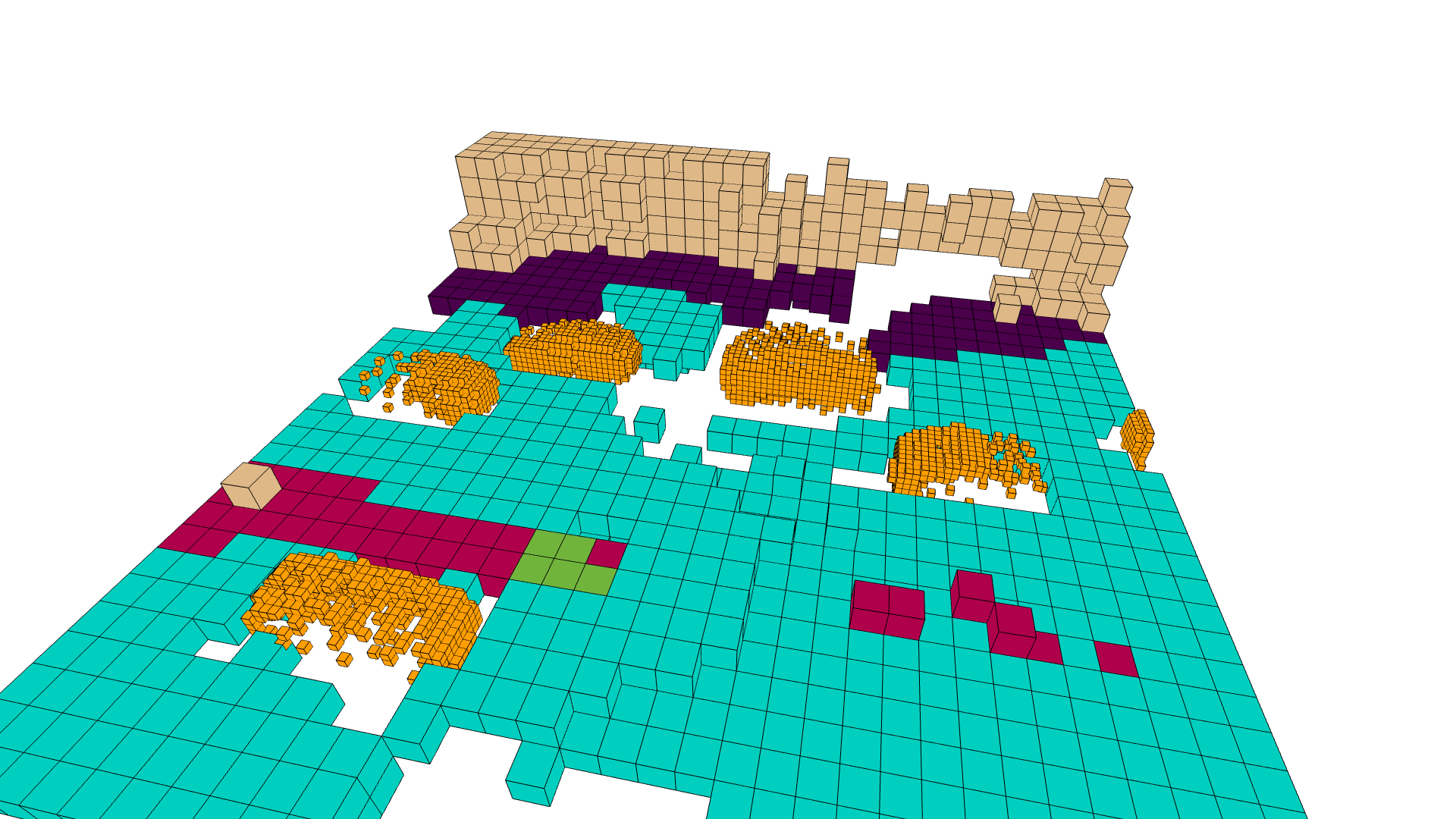}
    \includegraphics[width=0.32\textwidth]{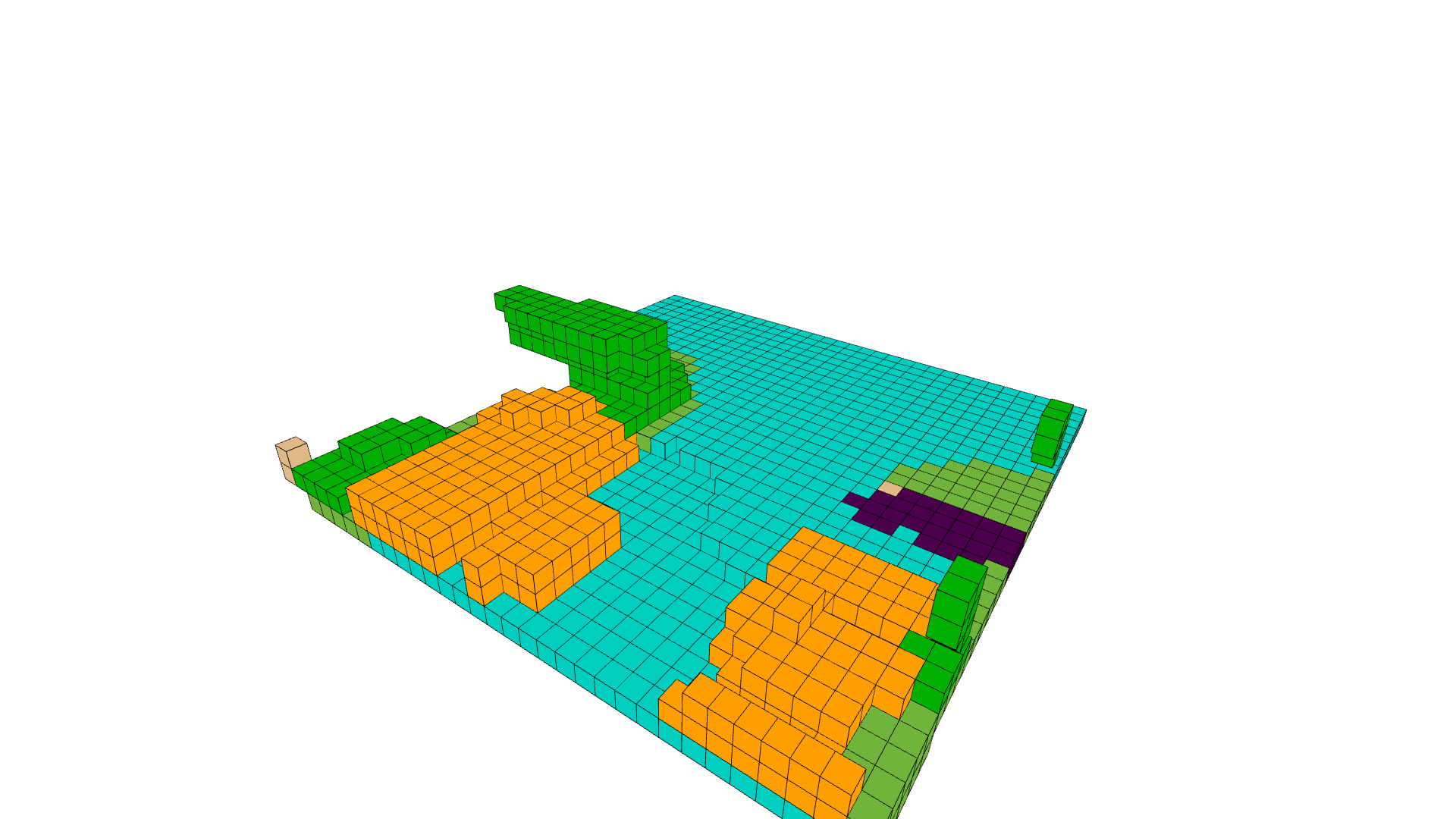}
    \includegraphics[width=0.32\textwidth]{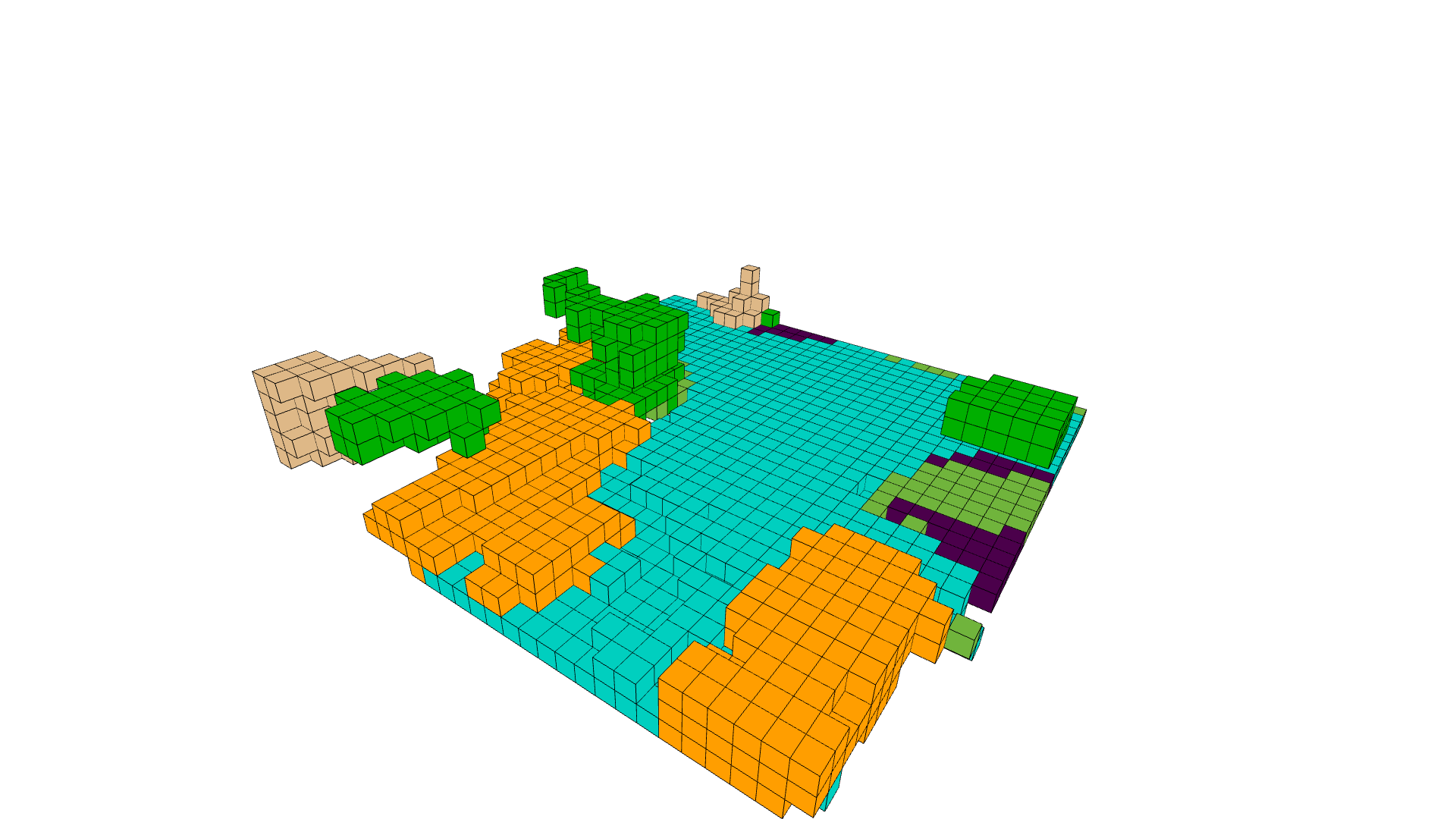}
    \includegraphics[width=0.32\textwidth]{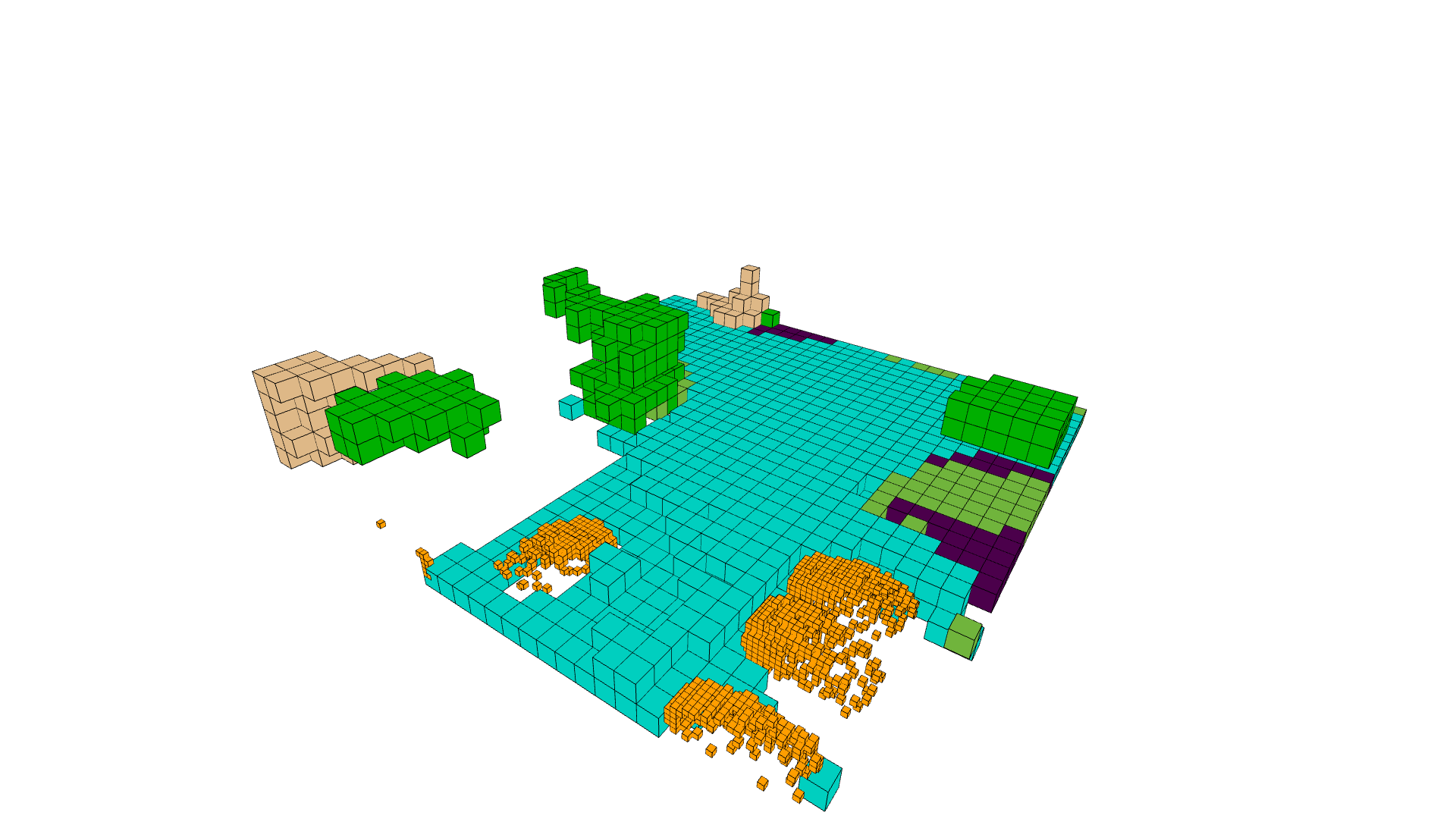}
    \includegraphics[width=0.32\textwidth]{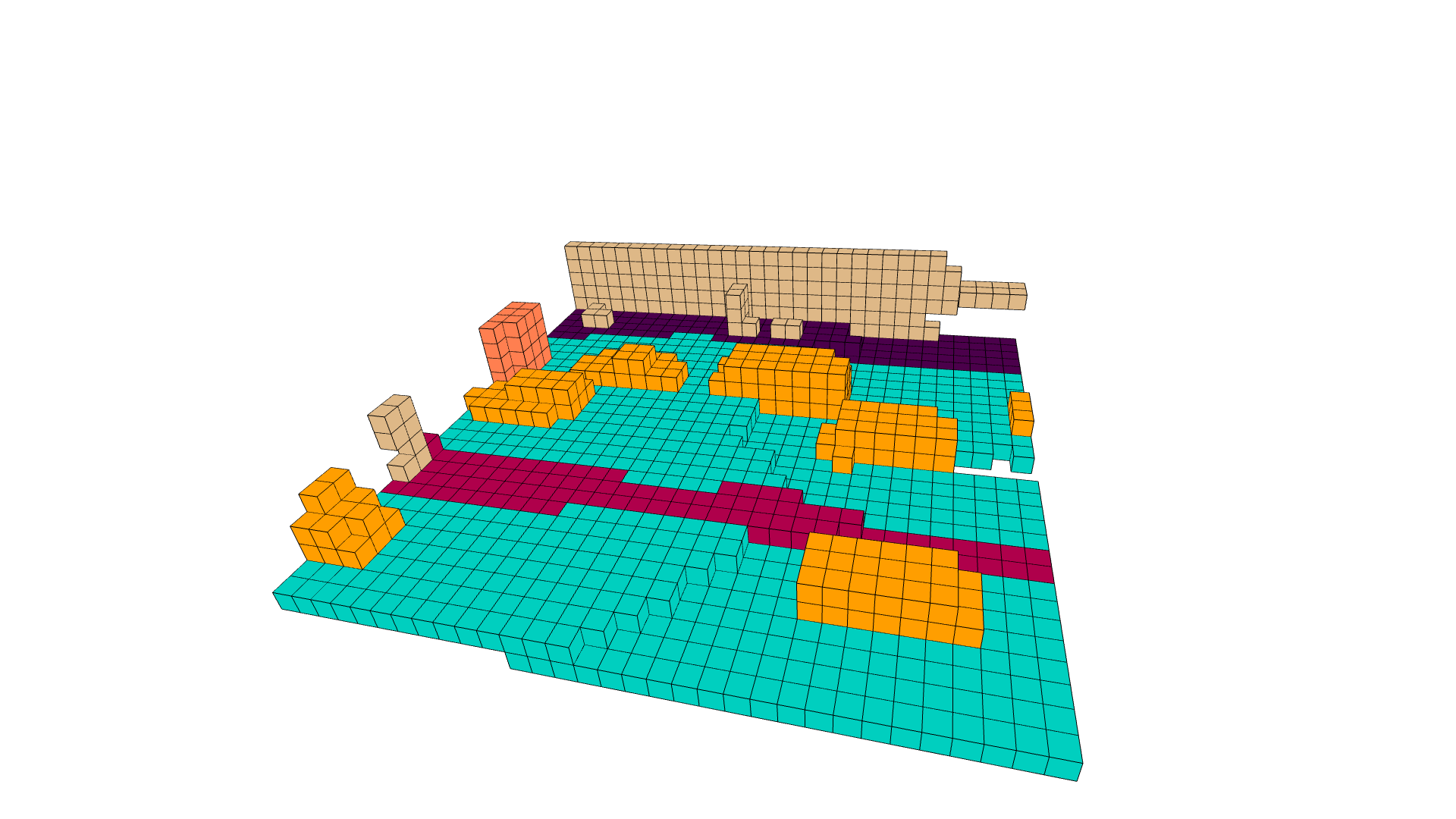}
    \includegraphics[width=0.32\textwidth]{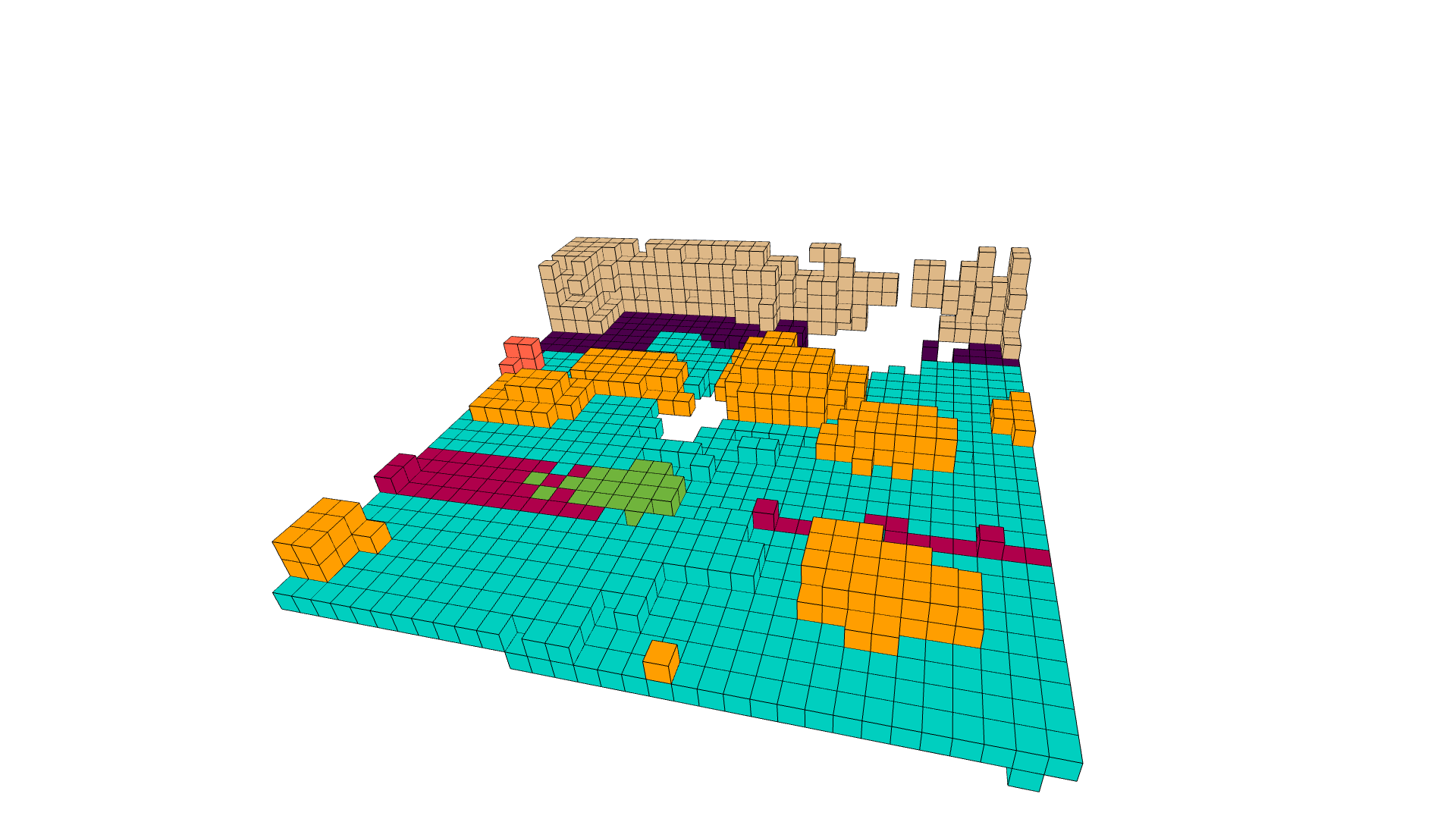}
    \includegraphics[width=0.32\textwidth]{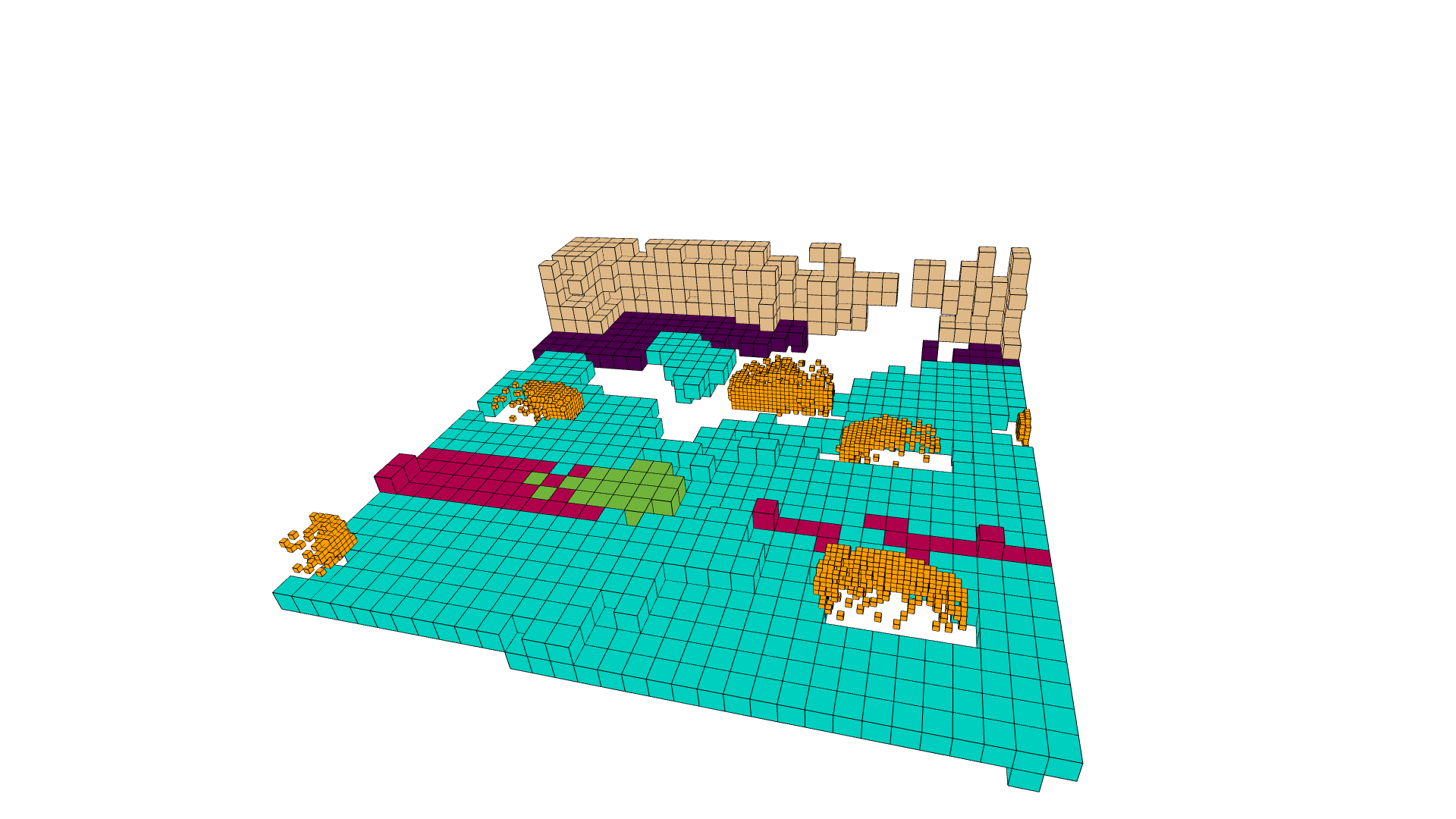}
    \includegraphics[width=0.32\textwidth]{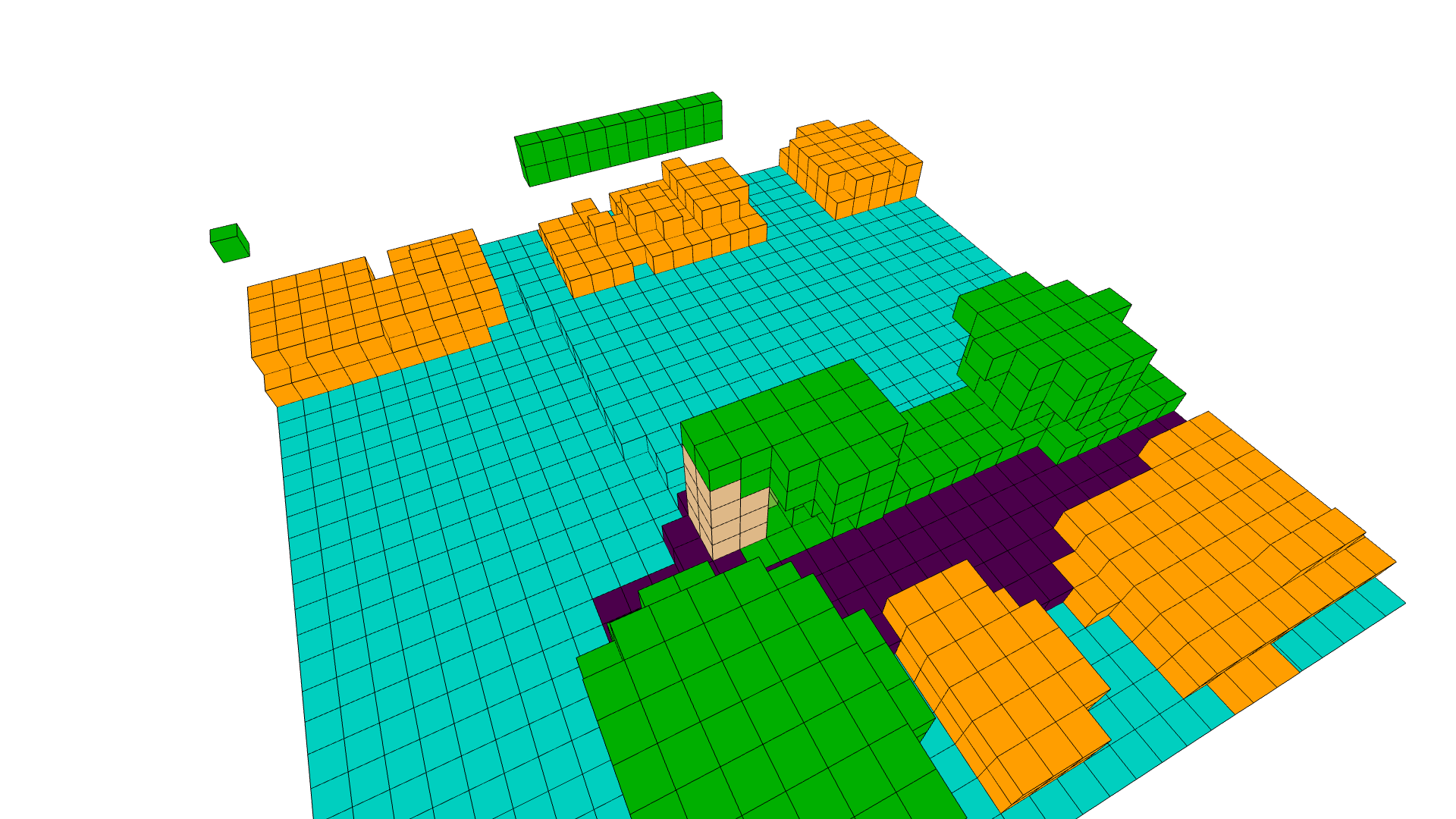}
    \includegraphics[width=0.32\textwidth]{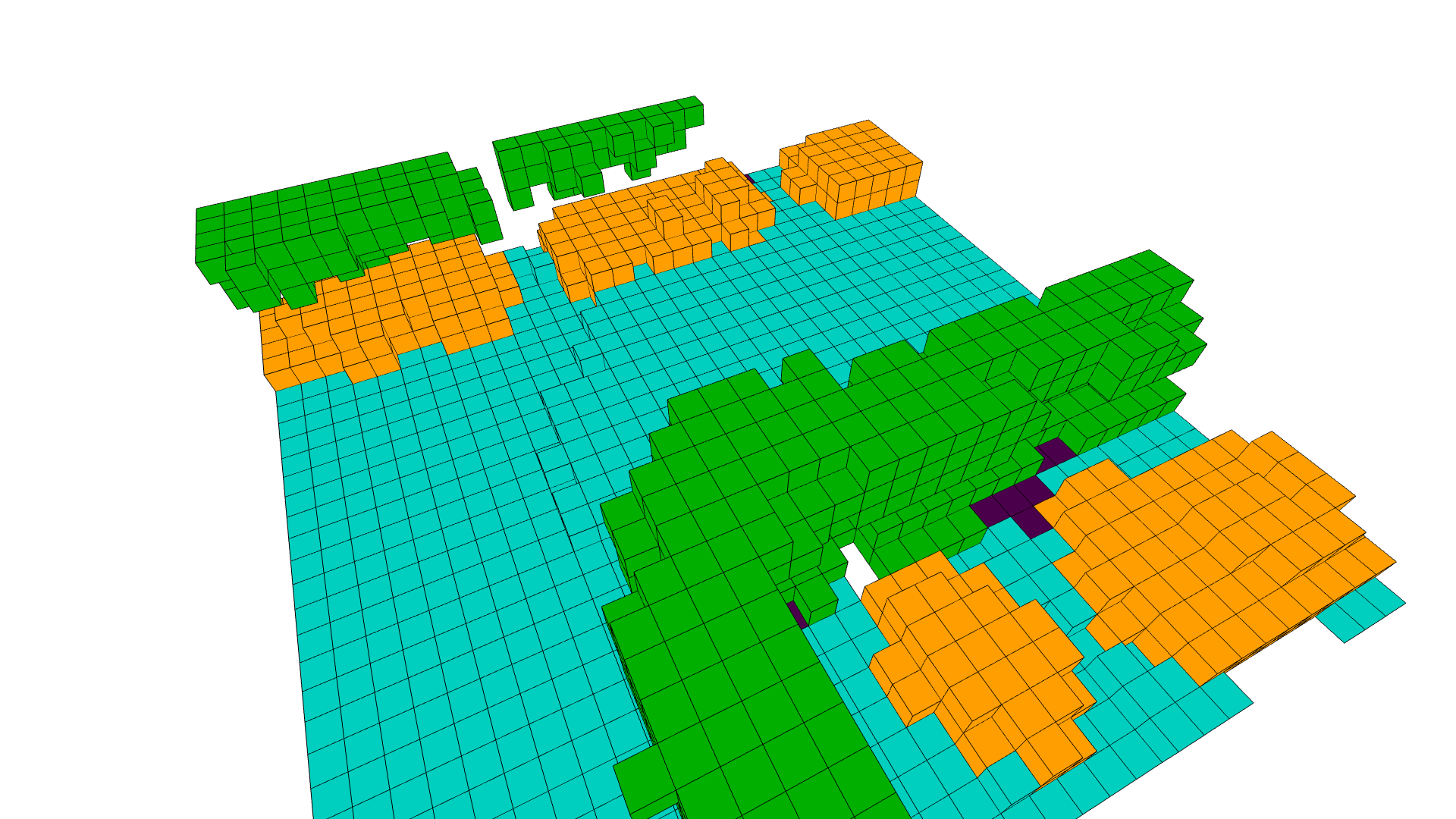}
    \includegraphics[width=0.32\textwidth]{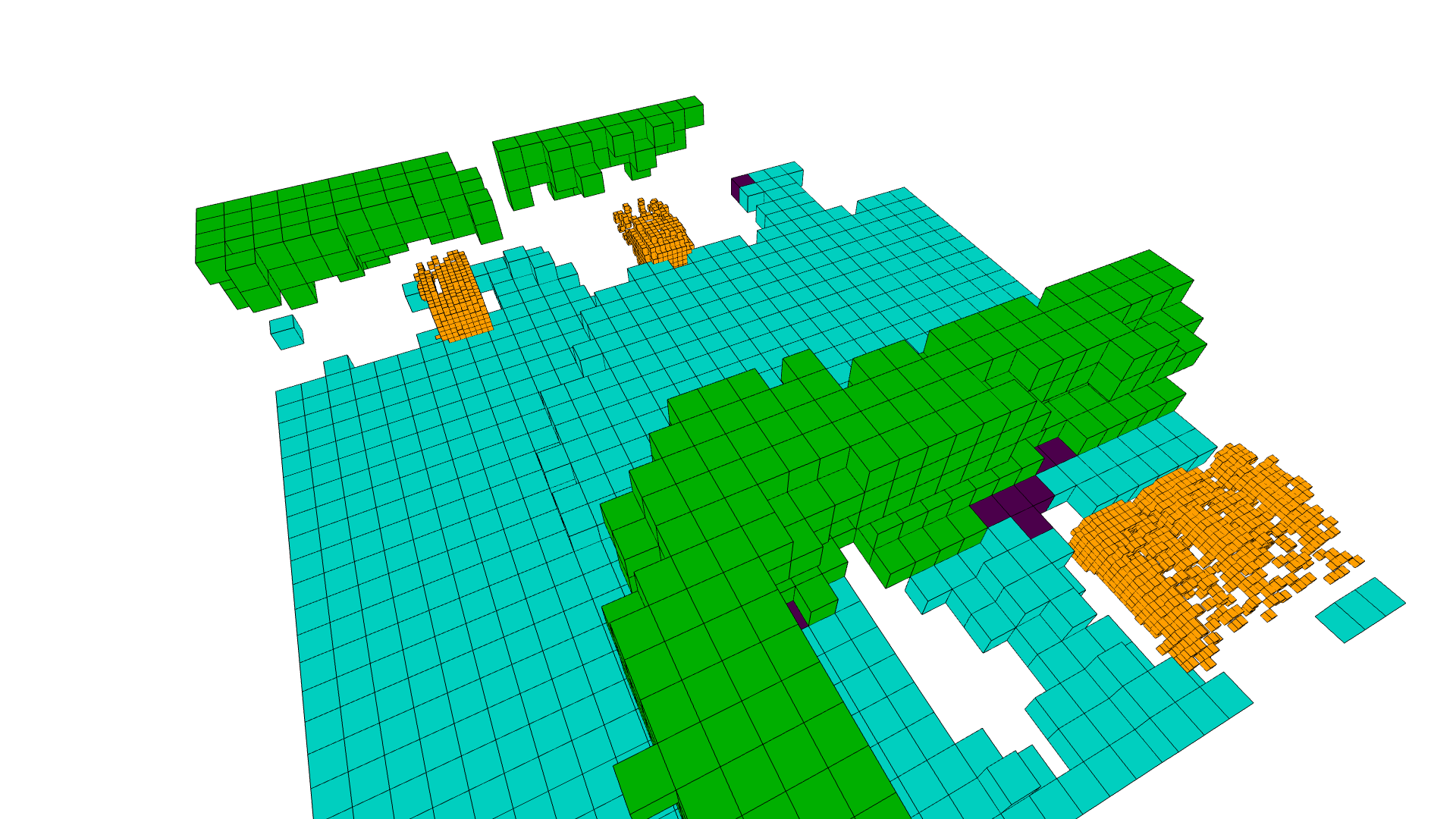}
    \caption{\textbf{Visualization on BEVFormer vs CONet vs. \titlevariable~(We visualize the pointcloud in a voxelized format)}.}
    \label{fig:coarse_fine_images_more}
\end{figure*}

\end{document}